\pdfoutput=1
\documentclass[11pt]{article}

\usepackage{acl}

\usepackage{times}
\usepackage{latexsym}

\usepackage[T1]{fontenc}
\usepackage[utf8]{inputenc}

\usepackage{microtype}

\usepackage{hyperref}
\usepackage{url}
\usepackage{amsmath,amsthm,amsfonts,amssymb,amscd}
\usepackage{mathrsfs}
\usepackage{times}
\usepackage{graphicx} % 插入图像
\usepackage{wrapfig}
\usepackage{subcaption}
\usepackage{multirow}
\usepackage{arydshln}
\usepackage{booktabs}
\usepackage{array}
\usepackage{rotating}
\usepackage{annotate-equations}

\title{Unified Interpretation of Smoothing Methods for Negative Sampling Loss Functions in Knowledge Graph Embedding}

\author{
  Xincan Feng\textsuperscript{\dag}, Hidetaka Kamigaito\textsuperscript{\dag}, Katsuhiko Hayashi\textsuperscript{\ddag}, Taro Watanabe\textsuperscript{\dag} \\
  \textsuperscript{\dag}Nara Institute of Science and Technology \textsuperscript{\ddag}The University of Tokyo\\
  \texttt{\{feng.xincan.fy2, kamigaito.h, taro\}@is.naist.jp} \\
  \texttt{katsuhiko-hayashi@g.ecc.u-tokyo.ac.jp}}

\begin{document}
\maketitle
\begin{abstract}
Knowledge Graphs (KGs) are fundamental resources in knowledge-intensive tasks in NLP. Due to the limitation of manually creating KGs, KG Completion (KGC) has an important role in automatically completing KGs by scoring their links with KG Embedding (KGE). To handle many entities in training, KGE relies on Negative Sampling (NS) loss that can reduce the computational cost by sampling. Since the appearance frequencies for each link are at most one in KGs, sparsity is an essential and inevitable problem. The NS loss is no exception. As a solution, the NS loss in KGE relies on smoothing methods like Self-Adversarial Negative Sampling (SANS) and subsampling. However, it is uncertain what kind of smoothing method is suitable for this purpose due to the lack of theoretical understanding. This paper provides theoretical interpretations of the smoothing methods for the NS loss in KGE and induces a new NS loss, Triplet Adaptive Negative Sampling (TANS), that can cover the characteristics of the conventional smoothing methods. Experimental results of TransE, DistMult, ComplEx, RotatE, HAKE, and HousE on FB15k-237, WN18RR, and YAGO3-10 datasets and their sparser subsets show the soundness of our interpretation and performance improvement by our TANS.
\end{abstract}

\section{Introduction}

% KG, KGC, and KGE are what, KG, KGC, and KGE are important.
Knowledge Graphs (KGs)
%such as Freebase \citep{Bollacker2008FreebaseAC}, are verified tuple databases structured to
represent human knowledge using various entities and their relationships as graph structures. 
%Entities could be \textit{people}, \textit{objects}, \textit{locations}, and \textit{events}, relations could be \textit{is\_mother\_of}, \textit{has\_properties\_of}.
KGs are fundamental resources for knowledge-intensive tasks like dialog \citep{moon-etal-2019-opendialkg}, question answering \citep{KG-COVID-19}, named entity recognition \citep{liu2019kbert}, open-domain questions \citep{kg-reasoning-in-lm}, and recommendation systems \citep{gao2020deep}, etc. 
% \citet{liu2019kbert} inject domain knowledge from KGs into the pre-trained LM to solve the knowledge-driven problems. 
%In performing KG-related tasks, the sparsity of KGs is an important perspective \citep{neural-kg-reasoning}. However, it is not clear how to quantify sparsity in KGs properly. 

However, to create complete KGs, we need to consider a large number of entities and all their possible relationships. Taking into account the explosively large number of combinations between entities, only relying on manual approaches is unrealistic to make complete KGs. 

Knowledge Graph Completion (KGC) is a task to deal with this problem. KGC involves automatically completing missing links corresponding to relationships between entities in KGs. 
To complete the KGs, we need to score each link between entities.
For this purpose, current KGC commonly relies on Knowledge Graph Embedding (KGE) \citep{bordes2011learning}.
KGE models predict the missing relations, named link prediction, by learning structural representations.
In the current KGE, models need to complete a link (triplet) $(e_i,r_k,e_j)$ of entities $e_i$ and $e_j$, and their relationship $r_k$ by answering $e_i$ or $e_j$ from a given query $(?,r_k,e_j)$ or $(e_i,r_k,?)$, respectively.
Hence, KGE needs to handle a large number of entities and their relationships during its training.
%and are a classical approach for KGC.

%In addition, KGE models have high interpretability, thus, they can provide theoretical guidance for emerging research not restricted to those KG-related. \citet{zhang-etal-2020-pretrain} address the sparse and noisy annotations issue in KGs, and enrich the structural representations utilizing the textual representations provided via pre-trained language models (LMs). \citet{nayyeri2022integrating} integrate KGE and pre-trained LMs to enhance performance in link prediction tasks. \citet{unified} give a unified interpretation of softmax cross-entropy and negative sampling (NS) using KGE models as a case study. 

% NS is what, NS is useful
To handle a large number of entities and relationships in KGs, Negative Sampling (NS) loss \citep{ns} is frequently used for training KGE models. 
The original NS loss is proposed to approximate softmax cross-entropy loss to reduce computational costs by sampling false labels from its noise distribution in training.
\citet{complex} import the NS loss from word embedding to KGE with utilizing uniform distribution as its noise distribution.
%NS loss function essentially adjusts the sampling strategy to reduce the number of negative examples and introduce diverse negative examples, making the sample distribution more closely reflect the true distribution. This helps address problems such as data sparsity and model robustness. Therefore, appropriate NS can not only reduce computation but also improve the accuracy of the model.
%For the aforementioned functions and benefits, NS is widely applied in various machine learning and deep learning tasks that contain a large number of negative examples or have category imbalance problems, especially in fields such as natural language processing (NLP) and recommendation systems, e.g., word embedding \citep{ns}, language modeling (LM) \citep{melamud-etal-2017-simple}, contextualized embedding \citep{clark-etal-2020-pre, Clark2020ELECTRA:}. Recently, knowledge graph embedding (KGE) \citep{complex} models commonly use NS. Thus, we studied the NS and its variations by mainly focusing on KGE from theoretical and empirical aspects. 
%
% SANS is what, SANS is better in certain scenarios
\citet{rotate} extend the NS loss to Self-Adversarial Negative Sampling (SANS) loss for efficient training of KGE. Unlike the NS loss with uniform distribution, the SANS loss utilizes the training model's prediction as the noise distribution. Since the negative samples in the SANS loss become more difficult to discriminate for models in training, the SANS can extract models' potential compared with the NS loss with uniform distribution.

%This is done by adding a conditional probability term of a negative sample predicted by the training model in the loss. In comparison to NS loss, SANS gives more weight to the negative samples that the model finds challenging to predict given its current inference ability. This approach allows the model to better learn from these hard negative samples at each stage of its training. The improvement is reflected in the larger gradient brought by these negative samples, and the model's enhanced ability to select negative samples in subsequent stages. In essence, SANS encourages the model to focus more on "difficult" negative examples, which could lead to more informative updates and ultimately better model performance. SANS is shown to be more effective than traditional NS in certain scenarios. 

\begin{figure*}[t]
    \centering
    \includegraphics[width=.93\textwidth]{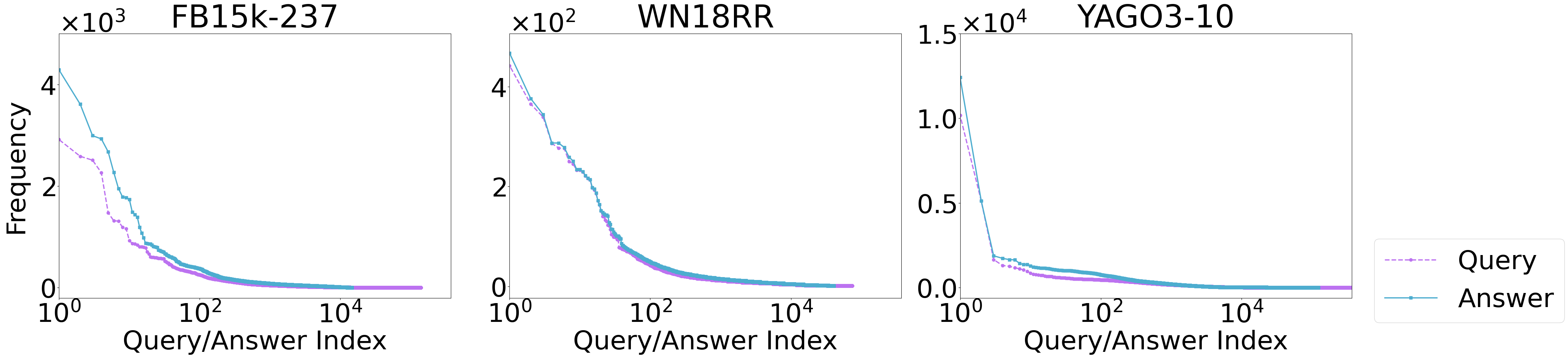}
    \caption{Appearance frequencies of queries and answers (entities) in the training data of FB15k-237, WN18RR, and YAGO3-10. Note that the indices are sorted from high frequency to low.}
    \label{fig:query_answer_freq}
\end{figure*}
\begin{figure*}[t]
    \centering
    \includegraphics[width=1\textwidth]{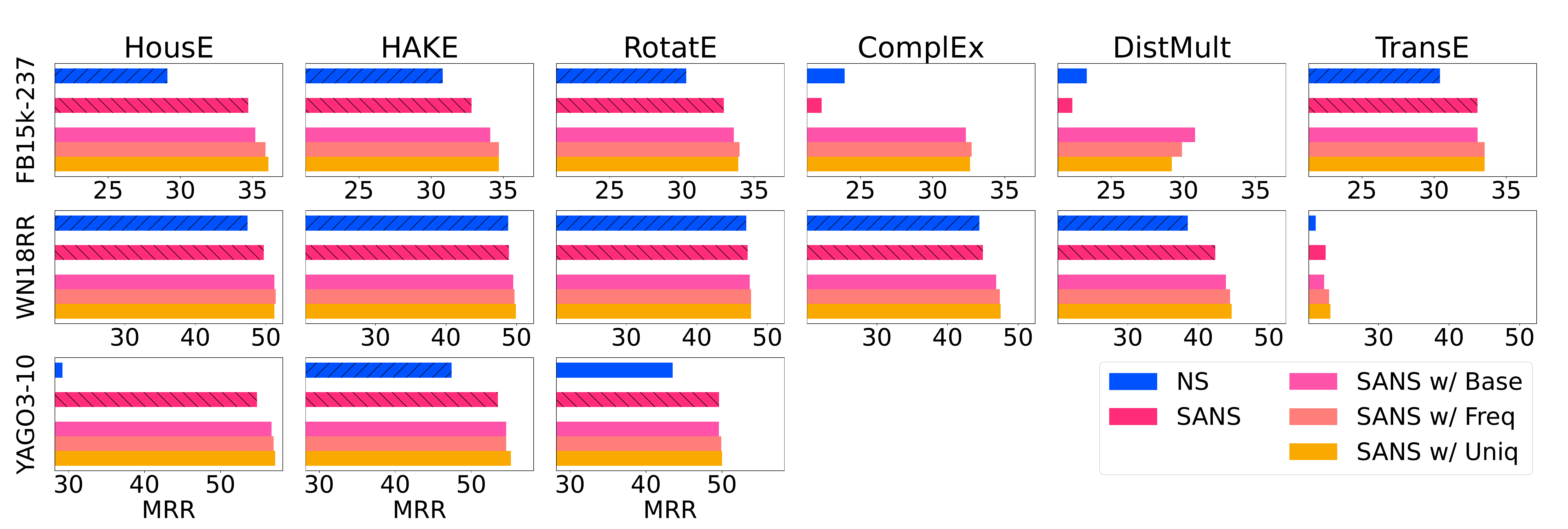}
    \caption{Performances of KGE models HousE, HAKE, RotatE, ComplEx, DistMult, and TransE on datasets FB15k-237, WN18RR, and YAGO3-10 using NS, SANS, and subsampling methods (noted as \textit{Base, Freq, Uniq}).}
    \label{fig:intro_unified_loss}
\end{figure*}

%\input{inputs/unified_loss_functions}

% Subsampling is what, Subsampling is better in some scenarios but not always
One of the problems left for KGE is the sparsity of KGs. Figure \ref{fig:query_answer_freq} shows the appearance frequency of queries and answers (entities) in the training data of FB15k-237, WN18RR and YAGO3-10 datasets. From the long-tail distribution of this figure, we can understand that both queries and answers necessary for training KGE models may suffer from the sparsity problem.

As a solution, several smoothing methods are used in KGE. \citet{rotate} import subsampling from word2vec \citep{ns} to KGE. Subsampling can smooth the appearance frequency of triplets and queries in KGs. \citet{pmlr-v162-kamigaito22a} show a general formulation that covers the basic subsampling of \citet{rotate} (Base), their frequency-based subsampling (Freq) and unique-based subsampling (Uniq) for KGE.
\citet{unified} indicate that SANS has a similar effect of using label-smoothing \citep{szegedy2016rethinking} and thus SANS can smooth the frequencies of answers in training. Figure \ref{fig:intro_unified_loss} shows the effectiveness of SANS and subsampling in KGC performance. From the figure, since FB15k-237 is more sparse (imbalanced) than WN18RR and YAGO3-10 based on Figure \ref{fig:query_answer_freq}, we can understand that strategy in choosing smoothing methods have more considerable influences than models when data is sparse.

%Unlike the original NS and SANS that considers only the noise distribution, subsampling considers the distribution of both positive and negative samples. Because not only the negative samples but also the positive samples could be wrongly predicted. Subsampling is shown to be able to improve the KGE performance more than NS and SANS in certain situations. 

% The problem in SANS and subsampling
While SANS and subsampling can improve model performance by smoothing the appearance frequencies of triplets, queries, and answers, their theoretical relationship is not clear, leaving their capabilities and deficiencies a question. For example, conventional works \citep{rotate,zhang-etal-2020-pretrain,pmlr-v162-kamigaito22a}\footnote{Note that \citet{rotate,zhang-etal-2020-pretrain} use subsampling in their released implementation without referring to it in their paper.} jointly use SANS and subsampling with no theoretical background. Thus, there is a call for further interpretability and performance improvement.
%In addition, the frequencies of queries contained in the tuples might vary greatly, e.g., some queries only occur once, and some appear thousands of times. In such cases, current approaches are unreasonable because they don't consider the frequency of queries. 

% Our contribution
To solve the above problem, we theoretically and empirically study the differences of SANS and subsampling on three common datasets and their sparser subsets with six popular KGE models\footnote{Our code and data are available at \url{https://github.com/xincanfeng/ss_kge}.}. 
% To solve the above problem, we theoretically and empirically study the differences of SANS and subsampling on three common datasets and their sparser subsets with six popular KGE models. 
% We summarize our proposals and findings
%not restricted to KGE
% as follows:
Our contributions are as follows:
\begin{itemize}
    \item By focusing on the smoothing targets, we theoretically reveal the differences between SANS and subsampling and induce a new NS loss, Triplet Adaptive Negative Sampling (TANS), that can cover the smoothing target of both SANS and subsampling. 
    \item We theoretically show that TANS with subsampling can potentially cover the conventional usages of SANS and subsampling. 
    \item We empirically verify that TANS improves KGC performance on sparse KGs in terms of MRR.
    \item We empirically verify that TANS with subsampling can cover the conventional usages of SANS and subsampling in terms of MRR. 
    %\item Enhanced by our method, the superiority of the Uniq subsampling strategy for KGs is revealed empirically. This finding is possible to help reduce the method tuning costs when applying subsampling methods to a new KG. 
\end{itemize}

\section{Background}

In this section, we describe the problem formulation for solving KGC by KGE and explain the conventional NS loss functions in KGE.

\subsection{Formulation of KGE}

KGC is a research topic for automatically inferring new links in a KG that are likely but not yet known to be true.
To infer the new links by KGE, we decompose KGs into a set of triplets (links). By using entities $e_{i}$, $e_{j}$ and their relation $r_{k}$, we represent the triplet as $(e_{i},r_{k},e_{j})$. In a typical KGC task, a KGE model receives a query $(e_{i},r_{k},?)$ or $(?,r_{k},e_{j})$ and predicts the entity corresponding to $?$ as an answer.

%KGE is a well-known scalable approach for KGC.
In KGE, a KGE model scores a triplet $(e_{i},r_{k},e_{j})$ by using a scoring function $s_{\mathbf{\theta}}(x,y)$, where $\mathbf{\theta}$ denotes model parameters. Here, using a softmax function, we represent the existence probability $p_{\mathbf{\theta}}(y|x)$ for an answer $y$ of the query $x$ as follows:
\begin{equation}
    p_{\mathbf{\theta}}(y|x) = \frac{\exp(s_{\mathbf{\theta}}(x,y))}{\sum_{y' \in Y}\exp(s_{\mathbf{\theta}}(x,y'))},
    \label{eq:kge:softmax}
\end{equation}
where Y is a set of entities.

\subsection{NS Loss in KGE}
\label{sec:ns}

To train $s_{\mathbf{\theta}}(x,y)$, we need to calculate losses for the observables $D=\{(x_{1},y_{1}),\cdots, (x_{n},y_{n})\}$ that follow $p_{d}(x,y)$.
Even if we can represent KGC by Eq.~(\ref{eq:kge:softmax}), it does not mean we can tractably perform KGC due to the large number of Y in KGs.
For the reason of the computational cost, the NS loss \citep{ns} is used to approximate Eq.~(\ref{eq:kge:softmax}) by sampling false answers.

By modifying that of \citet{ns}, the following NS loss \citep{rotate,ahrabian-etal-2020-structure} is commonly used in KGE:
\begin{align}
&\ell_{\text{NS}}(\mathbf{\theta}) \nonumber \\
= & -\frac{1}{|D|}\sum_{(x,y) \in D} \Bigl[\log(\sigma(s_{\mathbf{\theta}}(x,y)+\tau)) \nonumber \\
& + \frac{1}{\nu}\sum_{y_{i}\sim U}^{\nu}\log(\sigma(-s_{\theta}(x,y_i)-\tau))\Bigr],\label{eq:ns}
\end{align}
where $U$ is the noise distribution that follows uniform distribution, $\sigma$ is the sigmoid function, $\nu$ is the number of negative samples per positive sample $(x,y)$, and $\tau$ is a margin term to adjust the value range decided by $s_{\mathbf{\theta}}(x,y)$.

\subsection{Smoothing Methods for the NS Loss in KGE}
\label{subsec:smoothing_kge}

As shown in Figure \ref{fig:query_answer_freq}, KGC needs to deal with the sparsity problem caused by low frequent queries and answers in KGs. Imposing smoothing on the appearance frequencies of queries and answers can mitigate this problem. The following subsections introduce subsampling \citep{ns,rotate,pmlr-v162-kamigaito22a} and SANS \citep{rotate}, the conventional smoothing methods for the NS loss in KGE. 

\subsubsection{Subsampling}
\label{sec:subsampling}

%Equation~(\ref{eq:ns}) is on the assumption that the NS loss function fits the model to the distribution $p_{d}(y|x)$ defined from the observed data. However, what the NS loss actually does is to fit the model to the true distribution $p'_{d}(y|x)$ that exists behind the observed data.
Subsampling \citep{ns} is a method to smooth the frequency of triplets or queries in the NS loss. \citet{rotate} import this approach from word embedding to KGE. \citet{icml2022erratum,pmlr-v162-kamigaito22a} add some variants to subsampling for KGC and  theoretically provide a unified expression of them as follows: 
%To fill in the gap between $p_{d}(y|x)$ and $p'_{d}(y|x)$, \citet{icml2022erratum,pmlr-v162-kamigaito22a} theoretically add $A(x,y)$ and $B(x)$ to Eq. (\ref{eq:ns}) as follows\footnote{We include the detailed derivation of this function in Appendix \ref{app:derivation}.}:
\begin{align}
&\ell_{\text{SUB}}(\mathbf{\theta}) \nonumber\\
=&-\frac{1}{|D|}\!\!\sum_{(x,y) \in D} \!\!\Bigl[A(x,y;\alpha)\log(\sigma(s_{\theta}(x,y)\!+\!\tau)) \! \nonumber\\
& +\!\frac{1}{\nu}\!\sum_{y_{i}\sim U}^{\nu}\!\!\!B(x,y;\alpha)\!\log(\sigma(\!-s_{\theta}(x,y_i)\!-\!\tau)\!)\!\Bigr],
\label{eq:ns:sub}
\end{align}
where $\alpha$ is a temperature term to adjust the frequecy of triplets and queries. Note that we incorporate $\alpha$ into Eq.~(\ref{eq:ns:sub}) to consider various loss functions even though \citet{icml2022erratum,pmlr-v162-kamigaito22a} do not consider $\alpha$. 
In this formulation, we can consider several assumptions for deciding $A(x,y;\alpha)$ and $B(x,y;\alpha)$.
%We discuss the assumptions in \S\ref{sec:Base:Freq:Uniq} and \S\ref{sec:assumption issue}.
We introduce these assumptions in the following paragraphs:

%\subsubsection{Subsampling Assumptions}
%\label{sec:Base:Freq:Uniq}

\paragraph{Base}
As a basic subsampling approach, \citet{rotate} import the one originally used in word2vec \cite{ns} to KGE, defined as follows:
\begin{equation}
 A(x,y;\alpha)\!=\!B(x,y;\alpha)\!=\!\frac{\#(x,y)^{-\alpha}|D|}{\sum_{(x'\!\!,y') \in D}\#(x',y')^{-\alpha}},
 \label{eq:subsamp:default}
\end{equation}
where $\#$ is the symbol for frequency and $\#(x,y)$ represents the frequency of $(x,y)$.
In word2vec, subsampling randomly discards a word by a probability $1-\sqrt{t/f}$, where $t$ is a constant value and $f$ is a frequency of a word. This is similar to randomly keeping a word with a probability $\sqrt{t/f}$. Thus, we can understand that Eq.~(\ref{eq:subsamp:default}) follows the original use in word2vec.
Since the actual $(x,y)$ occurs at most once in KGs, when $(x,y)=(e_{i},r_{k},e_{j})$, they approximate the frequency of $(x,y)$ as:
\begin{equation}
 \#(x,y) \approx \#(e_{i},r_{k})+\#(r_{k},e_{j}),
 \label{eq:subsamp:approx}
\end{equation}
based on the approximation of n-gram language modeling \citep{katz1987estimation}.
%where $(e_{i},r_{k})$ is the tail-batch query that aims to predict the tail $e_{j}$ given the head $e_{i}$ and relation $r_{k}$, and $(r_{k},e_{j})$ is the head-batch query that aims to predict the head $e_{i}$ given the relation $r_{k}$ and tail $e_{j}$. 

%Different from the form in Equation~(\ref{eq:ns:sub}), Equation~(\ref{eq:subsamp:default}) uses $A(x,y)$ and $B(x,y)$ instead of $A(x,y)$ and $B(x)$. Thus, their approach does not follow the theoretically induced loss function in Eq.~(\ref{eq:ns:sub}).

\paragraph{Freq}
\citet{pmlr-v162-kamigaito22a} propose frequency-based subsamping (Freq) by assuming a case that $(x,y)$ originally has a frequency, but the observed one in the KG is at most 1. 
% They define Freq as follows:
\begin{align}
 A(x,y;\alpha)&=\frac{\#(x,y)^{-\alpha}|D|}{\sum_{(x',y') \in D}\#(x',y')^{-\alpha}},\:\:\:\:\: \nonumber\\
 B(x,y;\alpha)&=\frac{\#x^{-\alpha}|D|}{\sum_{x' \in D}\#x'^{-\alpha}}.
 \label{eq:subsamp:freq}
\end{align}

\paragraph{Uniq}
\citet{pmlr-v162-kamigaito22a} also propose unique-based subsamping (Uniq) by assuming a case that the originally frequency and the observed one in the KG are both 1. 
% They define Uniq as follows:
\begin{equation}
 A(x,y;\alpha)=B(x,y;\alpha)=\frac{\#x^{-\alpha}|D|}{\sum_{x' \in D}\#x'^{-\alpha}}.
 \label{eq:subsamp:uniq}
\end{equation}

\subsubsection{SANS Loss}
\label{sec:sans}
SANS is originally proposed as a kind of NS loss to train KGE models efficiently by considering negative samples close to their corresponding positive ones. \citet{unified} show that using SANS is similar to imposing label-smoothing on Eq.~(\ref{eq:kge:softmax}). Thus, SANS is a method to smooth the frequency of answers in the NS loss.
The SANS loss is represented as follows:
\begin{align}
&\ell_{\text{SANS}}(\mathbf{\theta}) \nonumber\\
=&-\frac{1}{|D|}\sum_{(x,y) \in D} \Bigl[\log(\sigma(s_{\theta}(x,y)+\tau)) \nonumber\\
&+\!\! \sum_{y_{i}\sim U}^{\nu}p_{\theta}(y_i|x;\beta)\log(\sigma(\!-\!s_{\theta}(x,y_i)\!-\!\tau))\Bigr], \\
&p_{\theta}(y_i|x;\beta)\approx\frac{\exp(\beta s_{\theta}(x,y_i))}{\sum_{j=1}^{\nu}\exp(\beta s_{\theta}(x,y_j))},\label{eq:sans}
\end{align}
where $\beta$ is a temperature to adjust the distribution of negative sampling.
Different from subsampling, SANS uses $p_{\theta}(y_i|x;\beta)$ that is predicted by a model $\theta$ to adjust the frequency of the answer $y_i$. Since $p_{\theta}(y_i|x;\beta)$ is essentially a noise distribution, it does not receive any gradient during training.

\begin{table*}[t]
    \centering
    \small
    % \resizebox{0.75\textwidth}{!}{
    \begin{tabular}{cccccl}
    \toprule
        \multicolumn{2}{c}{\multirow{2.5}{*}{Method}} & \multicolumn{3}{c}{Smoothing} & \multicolumn{1}{c}{\multirow{2.5}{*}{Remarks}} \\
        \cmidrule{3-5}
         & & $p(x,y)$ & $p(y|x)$ & $p(x)$ & \\
         \midrule
         \multirow{3}{*}{Subsampling} & Base & $\checkmark$ & $\triangle$ & $\triangle$ & $p(y|x)$ and $p(x)$ are influenced by $p(x,y)$.\\
         & Uniq & $\triangle$ & $\times$ & $\checkmark$ & $p(x,y)$ is indirectly controlled by $p(x)$.\\
         & Freq & $\checkmark$ & $\triangle$ & $\checkmark$ & $p(y|x)$ is indirectly controlled by $p(x,y)$ or $p(x)$. \\
         \midrule
         \multicolumn{2}{c}{SANS} & $\triangle$ & $\checkmark$ & $\times$ & $p(x,y)$ is indirectly controlled by $p(y|x)$.\\
         \midrule
         \multicolumn{2}{c}{TANS} & $\checkmark$ & $\checkmark$ & $\checkmark$ & \\
    \bottomrule
    \end{tabular}
    % }
    \caption{The characteristics of each smoothing method for the NS loss in KGE (See \S\ref{subsec:smoothing_kge} for the details.) and our proposed TANS. $\checkmark$ and $\triangle$ respectively denote the method smooths the probability directly and indirectly. $\times$ denotes the method does not smooth the probability.}
    \label{tab:summary_smoothing_kge}
\end{table*}

\section{Triplet Adaptive Negative Sampling}

% In this section, we explain the details of our proposed Triplet Adaptive Negative Sampling (TANS). TANS is an extended version of SANS that can adjust the appearance frequencies of triplets, queries, and answers in training data, not covered by the conventional SANS.  

In this section, we explain our proposed Triplet Adaptive Negative Sampling (TANS) in detail. We first show the overview of our TANS through the comparison with the conventional smoothing methods of the NS loss for KGE (See \S\ref{subsec:smoothing_kge}) in \S\ref{subsec:TANS:overview} and after that we explain the details of TANS through its mathematical formulations in \S\ref{subsec:TANS:formulation} and \S\ref{subsec:TANS:interpretation}.

\subsection{Overview}
\label{subsec:TANS:overview}

TANS is fundamentally different from SANS, with SANS only taking into account the conditional probability of negative samples and TANS being a loss function that considers the joint probability of the pair of queries and their answers.

Table \ref{tab:summary_smoothing_kge} shows the characteristics of TANS and the conventional smoothing methods of the NS loss for KGE introduced in \S\ref{subsec:smoothing_kge}. These characteristics are based on the decomposition of $p_d(x,y)$, the appearance probability for the triplet $(x,y)$, into that of its answer $p_d(y|x)$ and query $p(x)$:
\begin{equation}
    p_d(x,y) = p_d(y|x)p_d(x)
    \label{eq:freq:decompose}
\end{equation}
In Eq.~(\ref{eq:freq:decompose}), smoothing both $p_d(y|x)$ and $p_d(x)$ is similar to smoothing $p_d(x,y)$. However, smoothing $p_d(x,y)$ does not ensure smoothing both $p_d(x)$ and $p_d(y|x)$ considering the case of only one of them being smoothed, and the left one being still sparse. Similarly, smoothing only $p_d(x)$ or $p_d(y|x)$ does not ensure $p_d(x,y)$ being smoothed due to the case where one of them is still sparse. In Table \ref{tab:summary_smoothing_kge}, we denote such a case where the method can influence the probability, but no guarantee of the probability be smoothed as $\triangle$.

In TANS, we aim to smooth $p_d(x,y)$ by smoothing both $p_d(y|x)$ and $p_d(x)$ based on Eq.~(\ref{eq:freq:decompose}).

\subsection{Formulation}
\label{subsec:TANS:formulation}

Here, we induce TANS from SANS with targeting to smooth $p_d(x,y)$ by smoothing both $p_d(y|x)$ and $p_d(x)$.
First, we assume a simple replacement from $p_{\theta}(y|x)$ to $p_{\theta}(x,y)$ in $\ell_{\text{SANS}}(\mathbf{\theta})$ of Eq.~(\ref{eq:sans}): 
\begin{align}
&-\frac{1}{|D|}\sum_{(x,y) \in D} \Bigl[\log(\sigma(s_{\theta}(x,y)+\tau)) \nonumber\\
&+ \sum_{y_{i}\sim U}^{\nu}p_{\theta}(x,y_i)\log(\sigma(-s_{\theta}(x,y_i)-\tau))\Bigr]. \label{eq:TANS:fail}
\end{align}
However, using Eq.~(\ref{eq:TANS:fail}) causes an imbalanced loss between the first and second terms since the sum of $p_{\theta}(x,y_i)$ on all negative samples is not always 1. Thus, Eq.~(\ref{eq:TANS:fail}) is impractical as a loss function.

As a solution, we focus on the decomposition $p_{\mathbf{\theta}}(x,y)=p_{\mathbf{\theta}}(y|x)p_{\mathbf{\theta}}(x)$ and the fact that the sum of $p_{\mathbf{\theta}}(y|x)$ of all negative samples is always 1. By using $p_{\mathbf{\theta}}(x)$ to make a balance between the first and second loss term, we can modify Eq.~(\ref{eq:TANS:fail}) and induce our TANS as follows:
\begin{align}
&\ell_{\text{TANS}}(\mathbf{\theta}) \nonumber\\
=& -\frac{1}{|D|}\sum_{(x,y) \in D} \!\!\!p_{\theta}(x;\gamma)\Bigl[\log(\sigma(s_{\theta}(x,y)+\tau)) \nonumber\\
&+\!\! \sum_{y_{i}\sim U}^{\nu}\!p_{\theta}(y_i|x;\beta)\log(\sigma(\!-s_{\theta}(x,y_i)\!\!-\!\!\tau))\!\Bigr], \label{eq:TANS}\\
&p_{\mathbf{\theta}}(x;\gamma) = \sum_{y_{i}\in D} p_{\mathbf{\theta}}(x,y_{i};\gamma),\:\:\:\:\:\nonumber\\
&p_{\mathbf{\theta}}(x,y_i;\gamma) \!\!=\!\! \frac{\exp{(\gamma s_{\theta}(x,y_i))}}{\sum_{(x',y')\in D}\!\exp{\!(\gamma s_{\theta}(x',y')\!)}}, \label{eq:TANS:pxy}
\end{align}
where $\gamma$ is a temperature to smooth the frequency of queries.
Since TANS uses a noise distribution decided by $p_\theta(x;\gamma)$ and $p_\theta(y_i|x;\beta)$, it does not propagate gradients through probabilities for negative samples, and thus, memory usage is not increased. 

\begin{table*}[t]
    \centering
    \small
    %\resizebox{0.95\textwidth}{!}{
    \begin{tabular}{cccl}
    \toprule
    \multicolumn{3}{c}{Temperature} & \multicolumn{1}{c}{\multirow{2}{*}{Induced NS Loss}} \\
    \cmidrule{1-3}
       $\alpha$ & $\beta$ & $\gamma$ &  \\
       \midrule
       $=0$ & $=0$ & $=0$ &  Equivalent to $\ell_{\text{NS}}(\mathbf{\theta})$, the basic NS loss in KGE (Eq.~(\ref{eq:ns})) \\
       $=0$ & $=0$ & $\neq 0$ & Currently does not exist  \\
       $=0$ & $\neq 0$ & $=0$ & Proportional to $\ell_{\text{SANS}}(\mathbf{\theta})$, the SANS loss (Eq.~(\ref{eq:sans})) \\
       $=0$ & $\neq 0$ & $\neq0$ & Equivalent to our $\ell_{\text{TANS}}(\mathbf{\theta})$, the TANS loss (Eq.~(\ref{eq:TANS})) \\
       $\neq 0$ & $=0$ & $=0$ & Proportional to $\ell_{\text{NS}}(\mathbf{\theta})$, the basic NS loss in KGE (Eq.~(\ref{eq:ns})) with subsampling in \S\ref{subsec:smoothing_kge}\\
       $\neq 0$ & $=0$ & $\neq 0$ & Currently does not exist \\
       $\neq 0$ & $\neq 0$ & $=0$ & Proportional to $\ell_{\text{SANS}}(\mathbf{\theta})$, the SANS loss (Eq.~(\ref{eq:sans})) with subsampling in \S\ref{subsec:smoothing_kge} \\
       $\neq 0$ & $\neq 0$ & $\neq 0$ & Equivalent to our $\ell_{\text{UNI}}(\mathbf{\theta})$, the unified NS loss in KGE (Eq.~(\ref{eq:unify})) \\
        &  &  & and also equivalent to our $\ell_{\text{TANS}}(\mathbf{\theta})$, the TANS loss (Eq.~(\ref{eq:TANS})) with subsampling in \S\ref{subsec:smoothing_kge} \\
    \bottomrule
    \end{tabular}
    %}
    \caption{The relationship among the loss functions from the viewpoint of the unified NS loss, $\ell_{\text{UNI}}(\mathbf{\theta})$ in Eq.~(\ref{eq:unify}).}
    \label{tab:unify}
\end{table*}

\subsection{Theoretical Interpretation}
\label{subsec:TANS:interpretation}

In this subsection, we discuss the difference and similarities among TANS and other smoothing methods for the NS loss in KGE. As shown in Table \ref{tab:summary_smoothing_kge}, the subsampling methods, Base and Freq, can smooth triplet frequencies similar to our TANS.
To investigate TANS from the view point of subsampling, we reformulate Eq.~(\ref{eq:TANS}) as follows:
% \begin{align}
% &\ell_{\text{TANS}}(\mathbf{\theta}) \nonumber\\
% = &-\frac{1}{|D|}\!\sum_{(x,y) \in D} \!\!\!\!\!A(x,y;\gamma)\Bigl[\log(\sigma(s_{\theta}(x,y)\!+\!\tau))\nonumber\\
% &+ \!\!\!\sum_{y_{i}\sim U}^{\nu}\!\!B(x,y;\beta,\gamma)\log(\sigma(-s_{\theta}(x,y_i)\!-\!\tau))\Bigr], \label{eq:TANS:reformulated}\\
% &A(x,y;\gamma) = p_{\theta}(x;\gamma), \:\:\:\:\:\nonumber\\
% &B(x,y;\beta,\gamma) = p_{\theta}(y_i|x;\beta)p_{\theta}(x;\gamma).
% \end{align}
\begin{align}
&\ell_{\text{TANS}}(\mathbf{\theta}) \nonumber\\
= &-\frac{1}{|D|}\!\sum_{(x,y) \in D} \!\!\!\!\!\Bigl[A(x,y;\gamma)\log(\sigma(s_{\theta}(x,y)\!+\!\tau))\nonumber\\
&+ \!\!\!\sum_{y_{i}\sim U}^{\nu}\!\!B(x,y;\beta,\gamma)\log(\sigma(-s_{\theta}(x,y_i)\!-\!\tau))\Bigr], \label{eq:TANS:reformulated}\\
&A(x,y;\gamma) = p_{\theta}(x;\gamma), \:\:\:\:\:\nonumber\\
&B(x,y;\beta,\gamma) = p_{\theta}(y_i|x;\beta)p_{\theta}(x;\gamma).
\end{align}
Apart from the temperature terms, $\alpha$, $\beta$, and $\gamma$, we can see that the general formulation of subsampling in Eq.~(\ref{eq:ns:sub}) and the above Eq.~(\ref{eq:TANS:reformulated}) has the same formulation. Thus, TANS is not merely an extension of SANS but also a novel subsampling method.

Even though their similar characteristic, TANS and subsampling have an essential difference: TANS smooths the frequencies by model-predicted distributions as in Eq.~(\ref{eq:TANS:pxy}), and the subsampling methods smooth them by counting appearance frequencies on the observed data as in Eq.~(\ref{eq:subsamp:default}), (\ref{eq:subsamp:approx}), (\ref{eq:subsamp:freq}), and (\ref{eq:subsamp:uniq}). For instance, TANS can work even when the entity or relations included in the target triplet appear more than once, which is theoretically different from conventional approaches.

Since the superiority of using either model-based or count-based frequencies depends on the model and dataset, we empirically investigate this point through our experiments.

\section{Unified Interpretation of SANS and Subsampling}
\label{sec:interpretation}

In the previous section, we understand that our TANS can smooth triplets, queries, and answers partially covered by SANS and subsampling methods. On the other hand, TANS only relies on model-predicted frequencies to smooth the frequencies. 
% \citet{neubig-dyer-2016-generalizing} point out the benefits of using count-based and model-predicted frequencies to complement each other in language modeling. 
\citet{neubig-dyer-2016-generalizing} point out the benefits of combining count-based and model-predicted frequencies in language modeling. This section integrates smoothing methods for the NS loss in KGE from a unified interpretation.

\subsection{Formulation}

We formulate the unified loss function by introducing subsampling (Eq.~(\ref{eq:ns:sub})) into our TANS (Eq.~(\ref{eq:TANS})) as follows:
\begin{align}
&\ell_{\text{UNI}}(\mathbf{\theta})\nonumber\\
=& \!-\!\!\frac{1}{|D|}\!\!\sum_{(x,y) \in D}\!\!\! p_{\theta}(x;\gamma)\!\Bigl[\!A(x,y;\alpha)\!\log(\sigma(s_{\theta}(x,y)\!+\!\tau)) \nonumber\\
&+ \!\!\eta\!\!\!\sum_{y_{i}\sim U}^{\nu}\!\!\!B(x,y;\alpha)p_{\theta}(y_i|x;\beta)\!\log(\sigma(\!-\!s_{\theta}(x,y_i)\!-\!\tau)\!)\!\Bigr]\!, \label{eq:unify}
\end{align}
where $\eta$ is a hyperparamter that can be any value to absorb the difference among the three different subsampling methods, Base, Uniq, and Freq. 

Here, we can induce the NS losses shown in our paper from Eq.~(\ref{eq:unify}) by changing the temperature parameters $\alpha$, $\beta$, and $\gamma$. Table \ref{tab:unify} shows the induced losses from our $\ell_{\text{UNI}}(\mathbf{\theta})$. Note that since $p_{\theta}(x;\gamma)$ only appears in our TANS, canceling $p_{\theta}(x;\gamma)$ by $\gamma=0$ induces an inequivalent but a proportional relationship to the conventional NS loss.

\subsection{Theoretical Interpretation}
\label{subsec:uni:interpretation}

As shown in Table \ref{tab:unify}, TANS w/ subsampling has characteristics of all smoothing methods for the NS loss in KGE introduced in this paper. Therefore, we can expect higher performance of TANS w/ subsampling than the combination of conventional methods, the basic NS, SANS, and subsampling. However, because TANS w/ subsampling uses subsampling in \S\ref{subsec:smoothing_kge}, we need to choose the one from Base, Uniq, and Freq for TANS w/ subsampling. Since this part is out of the scope of theoretical interpretation, we investigate this in the experiments.

\if0
\begin{figure*}[t]
    \centering
    \includegraphics[width=1\textwidth]{figures/unified_loss_functions_hbar2.pdf}
    \caption{Performances of KGE models on datasets FB15k-237, WN18RR using NS, SANS, T-SANS, and NS with subsampling (Notations are the same as in Figure \ref{fig:intro_unified_loss}).}
    \label{fig:results:tsans}
\end{figure*}

\begin{figure*}[t]
    \centering
    \includegraphics[width=1\textwidth]{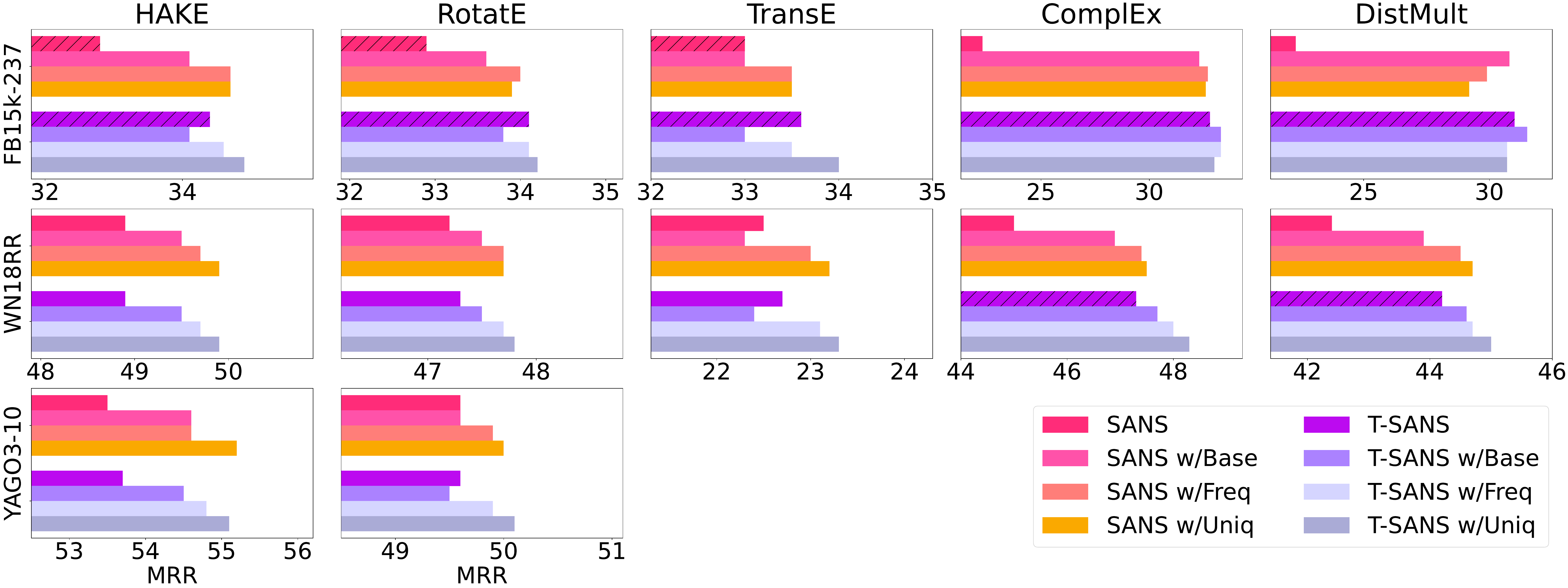}
    \caption{Performances of KGE models on datasets FB15k-237, WN18RR, and YAGO3-10 using SANS, T-SANS, and those with subsampling (Notations are the same as in Figure \ref{fig:intro_unified_loss}).}
    \label{fig:result_unified_loss}
\end{figure*}

\begin{figure*}[t]
    \centering
    \includegraphics[width=0.7\textwidth]{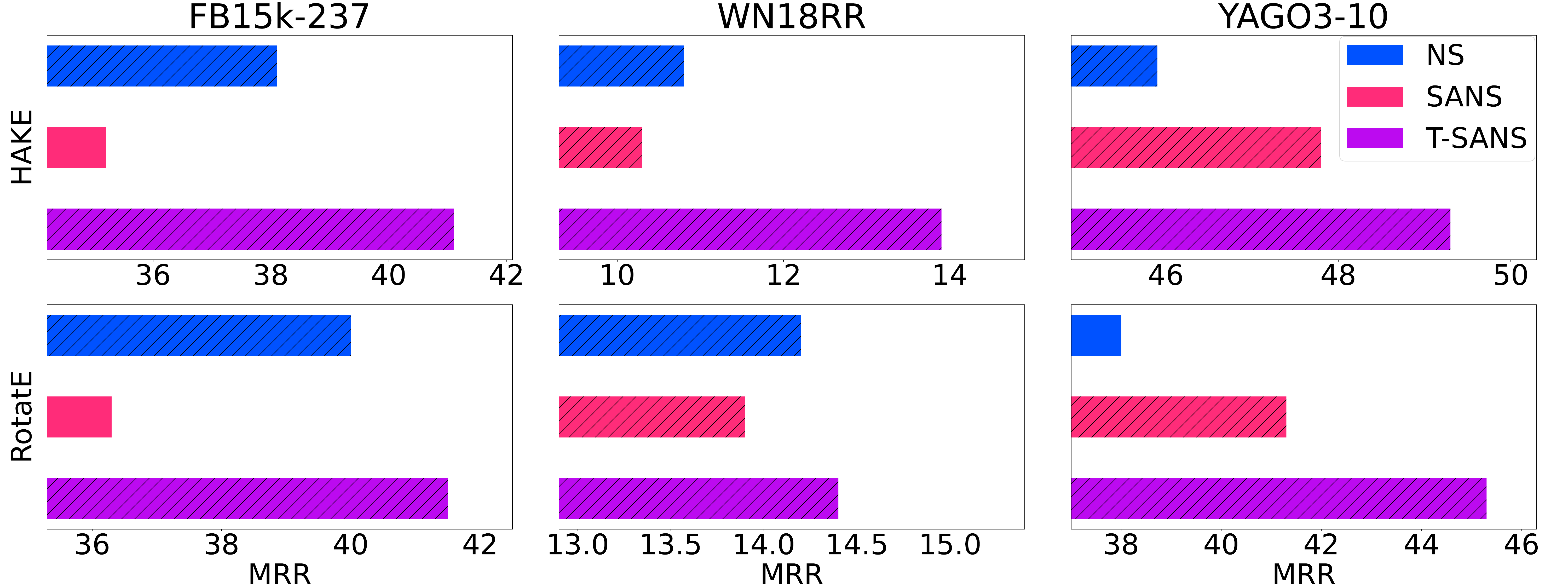}
    \caption{KGC performance on artificially created imbalanced KGs.}
    \label{fig:analysis_unified_loss}
\end{figure*}
\fi

\begin{figure*}[t]
\begin{subfigure}{\textwidth}
    \centering
    \includegraphics[width=1\textwidth]{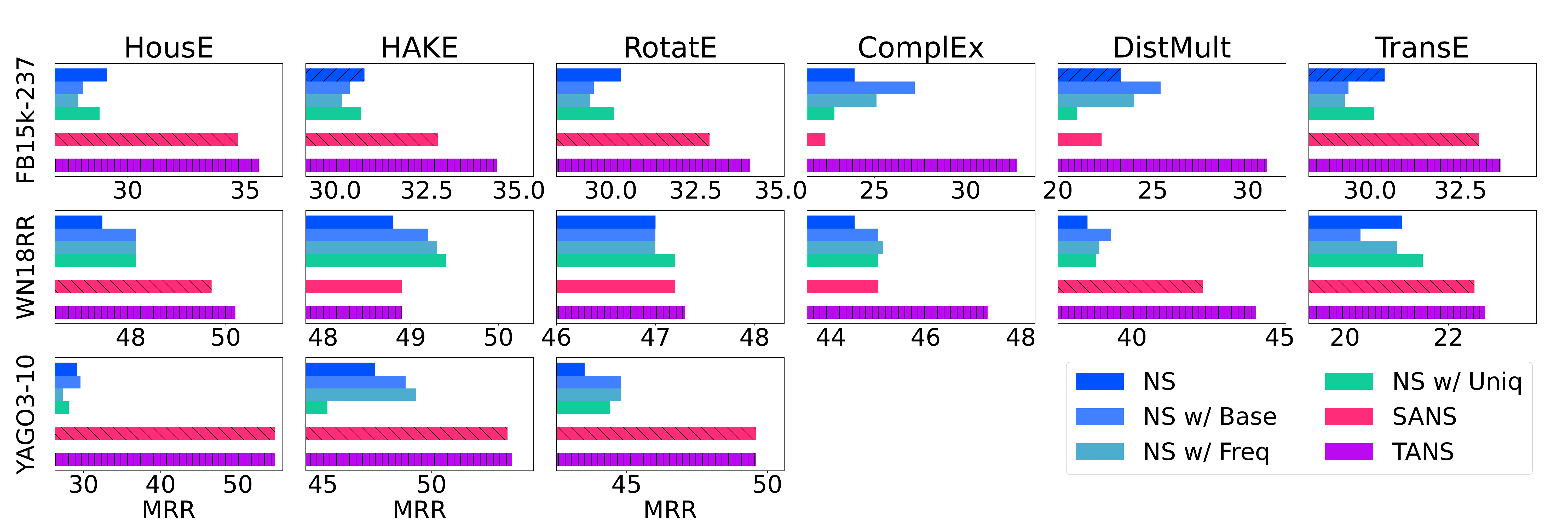}
    \caption{Results on datasets FB15k-237, WN18RR, YAGO3-10 using NS, SANS, TANS, and NS with subsampling.}
    \label{fig:results:tsans}
\end{subfigure}
\begin{subfigure}{\textwidth}
    \centering
    \includegraphics[width=1\textwidth]{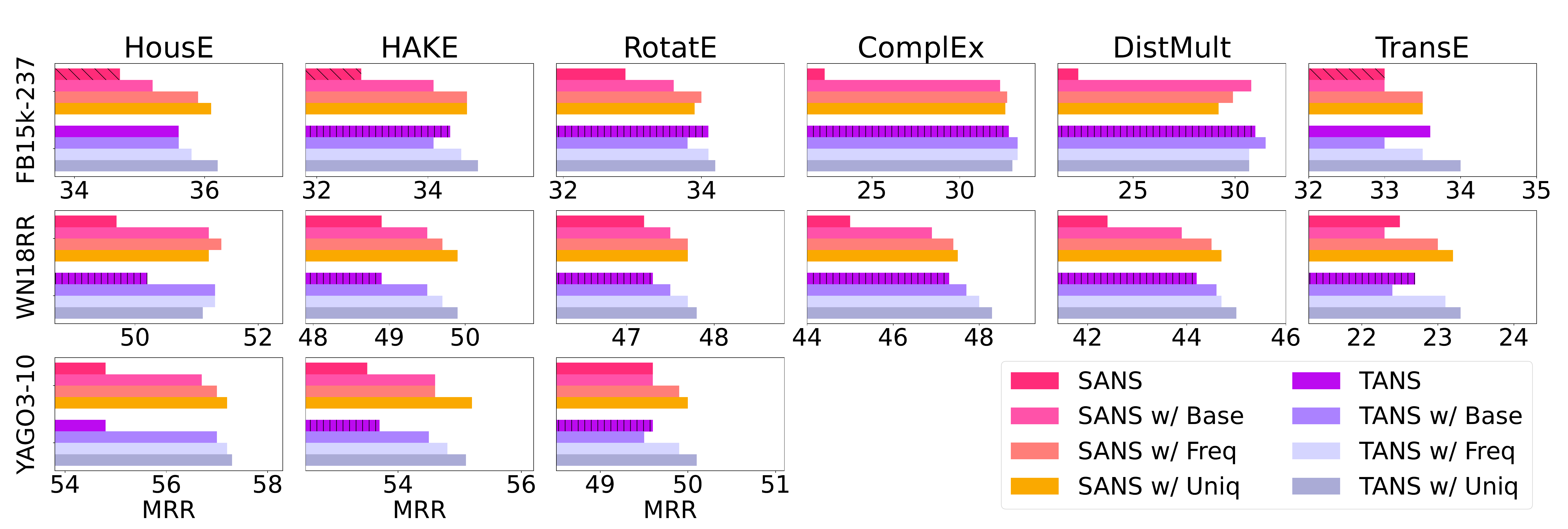}
    \caption{Results on datasets FB15k-237, WN18RR, YAGO3-10 using SANS, TANS, and those with subsampling.}
    \label{fig:result_unified_loss}
\end{subfigure}
\caption{KGC performance on common KGs (Notations are the same as in Figure \ref{fig:intro_unified_loss}).}
\end{figure*}

\begin{figure*}[t]
    \centering
    \includegraphics[width=1\textwidth]{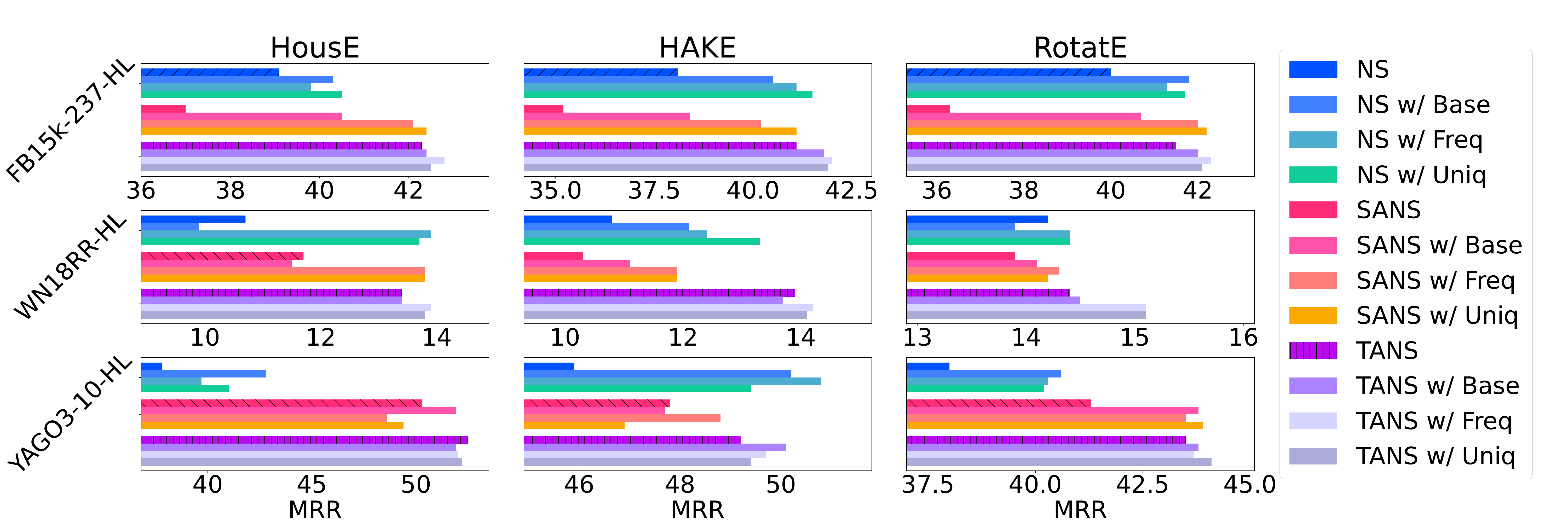}
    \caption{KGC performance on filtered sparser KGs, i.e., FB15k-237-HL, WN18RR-HL, and YAGO3-10-HL (Notations are the same as in Figure \ref{fig:intro_unified_loss}).}
    \label{fig:analysis_unified_loss}
\end{figure*}

\section{Experiments}

In this section, we investigate our theoretical interpretation in \S\ref{subsec:TANS:interpretation} and \S\ref{subsec:uni:interpretation} through experiments.

\subsection{Experimental Settings} 
\label{subsec:exp_setting}

\noindent\textbf{Datasets} We used three common datasets, FB15k-237~\citep{fb15k-237}, WN18RR, and YAGO3-10~\citep{conve}
% , and filtered their sparser subsets FB15k-237-HL, WN18RR-HL, and YAGO3-10-HL
\footnote{Table~\ref{tab:common_datasets}
% and \ref{tab:artificial_datasets} 
in Appendix \ref{app:sec:data_stats} shows the dataset statistics.}.

\noindent\textbf{Comparison Methods} As comparison methods, we used TransE~\citep{transe}, DistMult~\citep{distmult}, ComplEx~\citep{complex}, RotatE~\citep{rotate}, HAKE~\citep{hake}, and HousE~\citep{li2022house}. We followed the original settings of \citet{rotate} for TransE, DistMult, ComplEx, and RotatE with their implementation\footnote{\url{https://github.com/DeepGraphLearning/KnowledgeGraphEmbedding}}, the original settings of \citet{hake} for HAKE with their implementation\footnote{\url{https://github.com/MIRALab-USTC/KGE-HAKE}},  and the original settings of \citet{li2022house} for HousE with their implementation\footnote{\url{https://github.com/rui9812/HousE}}. We tuned temperature $\gamma$ on the validation split for each dataset.

\noindent\textbf{Metrics} We employed conventional metrics in KGC, i.e., MRR, Hits@1 (H@1), Hits@3 (H@3), and Hits@10 (H@10) and reported the average scores and their standard deviations by three different runs with fixed random seeds.

\subsection{Results}
\label{subsec:results}

Since the result tables are large\footnote{The full experimental results are listed in Appendix \ref{app:sec:full_exp}. The scores are included in Table~\ref{tab:fb15k-237},~\ref{tab:wn18rr},~and~\ref{tab:YAGO3-10} of Appendix \ref{app:sec:full_exp_table}. The training loss curves and validation MRR curves for each smoothing method are in Figure~\ref{fig:full_exp_train_valid_curves_fb},~\ref{fig:full_exp_train_valid_curves_wn},~and~\ref{fig:full_exp_train_valid_curves_yago} of Appendix \ref{app:sec:full_exp_fig}.}, we discuss them individually, focusing on important information in the following subsections.

\subsubsection{Effectiveness of TANS}

Figure~\ref{fig:results:tsans} shows the MRR scores of each method. From the result, we can understand the effectiveness of considering triplet information in SANS as conducted in TANS. Thus, the result is along with our expectation in \S\ref{subsec:TANS:interpretation} that TANS can cover the role of subsampling methods. However, as the result of HAKE on WN18RR shows, there is a case that subsampling methods outperform TANS. As discussed in \S\ref{subsec:TANS:interpretation}, using only TANS does not cover all combinations of NS loss and subsampling. Considering this theoretical fact, we further compare TANS with subsampling and the SANS loss with subsampling in the following section.

\subsubsection{Validity of the Unified Interpretation}

Figure~\ref{fig:result_unified_loss} shows the result for each configuration. We can see performance improvements by using subsampling in both SANS and TANS. Furthermore, in almost all cases, TANS with subsampling achieve the highest MRR. This observation is along with the theoretical conclusion in \S\ref{subsec:TANS:interpretation} that TANS with subsampling can cover the characteristic of other NS loss in terms of smoothing. On the other hand, the results of HAKE on YAGO3-10 show the different tendency that SANS with subsampling achieves the best MRR instead of TANS. Because the model prediction estimates the triplet frequencies, TANS is influenced by the selected model. Therefore, carefully choosing the combination of a loss function and model is still effective in improving KGC performance on the NS loss with subsampling.

\section{Analysis}
\label{sec:analysis}

%For this purpose, we extracted low-frequency triplets in two steps. First, we selected the highest and lowest 5\% frequent queries; then, the selected queries that have the highest and lowest 50\% frequent tuples are chosen. Tuples that contain the filtered queries consist of above artificial datasets.

We analyze how TANS mitigates the sparsity problem in imbalanced KGs commonly caused by low frequent triplets in KGC. By considering that all triplets in KGs appear at most once, we focus on queries. We extracted 0.5\% triplets with the highest or lowest frequent queries in training, validation, and test splits as the sparser subsets FB15k-237-HL, WN18RR-HL, and YAGO3-10-HL, respectively \footnote{Note that we show their appearance frequencies of queries and answers in the training data in Figure~\ref{fig:query_answer_freq_hl} and detailed statistics in Table~\ref{tab:artificial_datasets} of Appendix~\ref{fig:frequencies_hl} and ~\ref{app:subsec:sparse_stats}, respectively.} from original data, for the investigation. 
    
% Figure \ref{fig:analysis_unified_loss} shows MRRs for each experiment. From the result, we can understand that TANS can improve the KGC performance when KGs are sparse and imbalanced. You can see further details of the result in Table \ref{tab:artificial_datasets} of Appendix \ref{app:subsec:sparse_results}.

Figure~\ref{fig:analysis_unified_loss} shows MRRs for each model on each sparser dataset. From the result, we can understand that TANS can perform even much better in KGC when KGs get more imbalanced. You can see further detailed results in Table~\ref{tab:results_hl_fb}, ~\ref{tab:results_hl_wn}, and ~\ref{tab:results_hl_yago} of Appendix \ref{app:subsec:sparse_results}.

\section{Related Work}

\paragraph{Knowledge Graph} Knowledge graphs have important roles in various knowledge-intensive NLP tasks like dialog \citep{moon-etal-2019-opendialkg}, question answering \citep{KG-COVID-19}, named entity recognition \citep{liu2019kbert}, open-domain questions \citep{kg-reasoning-in-lm}, recommendation systems \citep{gao2020deep}, and commonsense reasoning \citep{sakai2024mcsqamultilingualcommonsensereasoning}, etc. In addition to these text-only tasks, knowledge-intensive vision and language (V\&L) tasks such as visual question answering (VQA) \citep{yue2023mmmu}, image generation \citep{kamigaito-etal-2023-table}, explanation generation \citep{hayashi2024artwork}, and image review generation \citep{saito2024evaluating} also require external knowledge. Visual KGs \citep{9961954} have the potential to contribute to solving these tasks. Therefore, KGs are important materials in various different fields. 

\paragraph{Knowlege Graph Completion}
Even though KGs are useful, their sparsity is a fundamental problem. To solve the sparsity of knowledge graphs, we need to complete them by inferring their unseen links between nodes, which are entities. For that purpose, knowledge graph completion (KGC) and knowledge graph embedding (KGE) \cite{bordes2011learning}, which represents KG information as a continuous vector space, are commonly used. As KGE methods, vector space models like TransE~\cite{transe}, DistMult~\citep{distmult}, ComplEx~\citep{complex}, RotatE~\citep{rotate}, HAKE~\citep{hake}, and HousE~\citep{li2022house}, that learn only from task-specific datasets expand this field as pioneers. As well as such approaches, pre-trained language model (PLM)-based approaches like KEPLER~\cite{wang-etal-2021-kepler} and SimKGC~\cite{wang-etal-2022-simkgc} also have an important role in KGC due to their ability to utilize the knowledge obtained in pre-training. However, as pointed out by~\citet{sakai-etal-2024-pre}, PLM-based approaches have a leakage issue caused by data contamination in pre-training.
Generation-based KGC methods like KGT5 \cite{saxena2022kgt5} and GenKGC \cite{10.1145/3487553.3524238} are unique in directly generating entity names. In hierarchical text classification (HTC), generation-based approaches contribute to improving performance \cite{kwon-etal-2023-hierarchical} supported by considering label hierarchies by fusing pre-trained text and label embeddings \cite{xiong-etal-2021-fusing,zhang-etal-2021-language} on the decoder.
However, \citet{sakai-etal-2024-pre} point out that commonly used KGC methods conduct link-level prediction, and such generation-based KGC methods make it difficult to use structure information of KGs directly. Thus, their performance gain is limited.
This situation requires investigating the benefits of inferring links by generation-based KGC under predefined entities and relationships.

\paragraph{Negative Sampling}

\citet{ns} initially propose the NS loss of the frequent words to train their word embedding model, word2vec. \citet{complex} introduce the NS loss to KGE to speed up training. \citet{melamud-etal-2017-simple} use the NS loss to train the language model. In contextualized pre-trained embeddings, \citet{clark-etal-2020-pre} indicate that a BERT \citep{devlin-etal-2019-bert}-like model ELECTRA \citep{Clark2020ELECTRA:} uses the NS loss to perform better and faster than language models. \citet{rotate} extend the NS loss to SANS loss for KGE and propose their noise distribution, which is subsampled by a uniformed probability $p_\theta(y_i|x)$. \citet{unified} point out the sparseness problem of KGs through their theoretical analysis of the NS loss in KGE. Furthermore, \citet{pmlr-v162-kamigaito22a,icml2022erratum} reveal that subsampling \cite{ns} can alleviate the sparseness problem in the NS for KGE and conclude three assumptions for subsampling, i.e., Base, Freq, and Uniq. \citet{feng-etal-2023-model} incorporate their proposed model-based subsampling that estimates frequencies for entities and their relationships by a trained KGE model into the subsampling of the NS loss to mitigate the sparseness issue of counting the frequency by increasing computational cost to train the additional KGE model.

\paragraph{Our Work}
Through our work, we theoretically clarify the position of the previous works on SANS loss and subsampling from the viewpoint of smoothing methods for the NS loss in KGE. Since our work unitedly interprets SANS loss and subsampling, our proposed TANS inherits the advantages of conventional works and can deal with the sparsity problem in the NS loss for KGE.

\section{Conclusion}

We reveal the relationships between SANS loss and subsampling for the KG completion task through theoretical analysis. We explain that SANS loss and subsampling under three assumptions, Base, Freq, and Uniq have similar roles to mitigate the sparseness problem of queries and answers of KGs by smoothing the frequencies of queries and answers. Furthermore, based on our interpretation, we induce a new loss function, Triplet Adaptive Negative Sampling (TANS), by integrating SANS loss and subsampling. We also introduce a theoretical interpretation that TANS with subsampling can cover all conventional combinations of SANS loss and subsampling.

We verified our interpretation by empirical experiments in three common datasets, FB15k-237, WN18RR, and YAGO3-10, and six popular KGE models, TransE, DistMult, ComplEx, RotatE, HAKE, and HousE. The experimental results show that our TANS loss can outperform subsampling and SANS loss with many models in terms of MRR as expected by our theoretical interpretation. Furthermore, the combinatorial use of TANS and subsampling achieved comparable or better performance than other combinations and showed the validity of our theoretical interpretation that TANS with subsampling can cover all conventional combinations of SANS loss and subsampling in KGE.

%In our future work, we plan to generalize TANS for word embeddings and item recommendations tasks, since these are similar to the special case of KGs whose triplets have the same relationships.

% \section{Future Work}
%We will investigate the properties of loss functions if there are larger and noisier datasets. 
% In addition, we will generalize our method in deep learning models and measure validity in the transformer model and language models like BERT and GPT. 
%Besides, because the negative samples are randomly chosen, we hope to investigate if our subsampling loss could be applied in quantum computing. We can try leveraging the state of superposition to process multiple negative samples concurrently in a single operation.
%In addition, by using the NS loss and its variations, the model can learn which samples are likely to be mispredicted. Therefore, we want to design a new machine-learning architecture that can feed these hard-to-predict samples back into the sampling process, replacing the randomly chosen negative samples. We can let the model learn easy-to-predict samples first, and gradually increase the difficulty to encourage the model to learn the hard-to-predict samples. 

\section*{Limitations}

Our experiments are conducted exclusively on public datasets, which are relatively well-balanced. Consequently, we anticipate that our TANS will perform better on real-world KGs.

\section*{Ethics Statement}

We used the publicly available datasets, FB15k-237, WN18RR, and YAGO3-10, to train and evaluate KGE models, and there is no ethical consideration.

\section*{Reproducibility Statement}

We used the publicly available code to implement KGE models, TransE, DistMult, ComplEx, RotatE, HAKE, and HousE with the author-provided hyperparameters as described in \S\ref{subsec:exp_setting}. Regarding the temperature parameter $\gamma$, we tuned it on the validation split for each dataset and reported the values in Table~\ref{tab:fb15k-237},~\ref{tab:wn18rr},~and~\ref{tab:YAGO3-10} of Appendix \ref{app:sec:full_exp}. Our code and data are available at \url{https://github.com/xincanfeng/ss_kge}.

% \section*{Acknowledgements}
\section*{Acknowledgements}
This work was supported by NAIST Granite, i.e., JST SPRING Grant Number JPMJSP2140.

\bibliography{anthology,custom}
\bibliographystyle{acl_natbib}

\appendix

\section{Dataset Statistics}
\label{app:sec:data_stats}
Table~\ref{tab:common_datasets} shows the dataset statistics for dataset FB15k-237, WN18RR, and YAGO3-10, introduced in \S\ref{subsec:exp_setting}.

\section{Full Experimental Results}
\label{app:sec:full_exp}

\subsection{Results Tables}
\label{app:sec:full_exp_table}
Table~\ref{tab:fb15k-237},~\ref{tab:wn18rr},~and~\ref{tab:YAGO3-10} list all results on FB15k-237, WN18RR, and YAGO3-10, explained in \S\ref{subsec:results}.
In these tables, the bold scores are the best results for each subsampling type (e.g. \textit{None}, \textit{Base}, \textit{Freq}, and \textit{Uniq}.), $\dagger$ indicates the best scores for each model, \textit{SD} denotes the standard deviation of the three trials, and $\gamma$ denotes the temperature chosen by development data.

\subsection{Training Loss and Validation MRR Curve}
\label{app:sec:full_exp_fig}
\begin{figure*}[t]
    \centering
    \includegraphics[width=\textwidth]{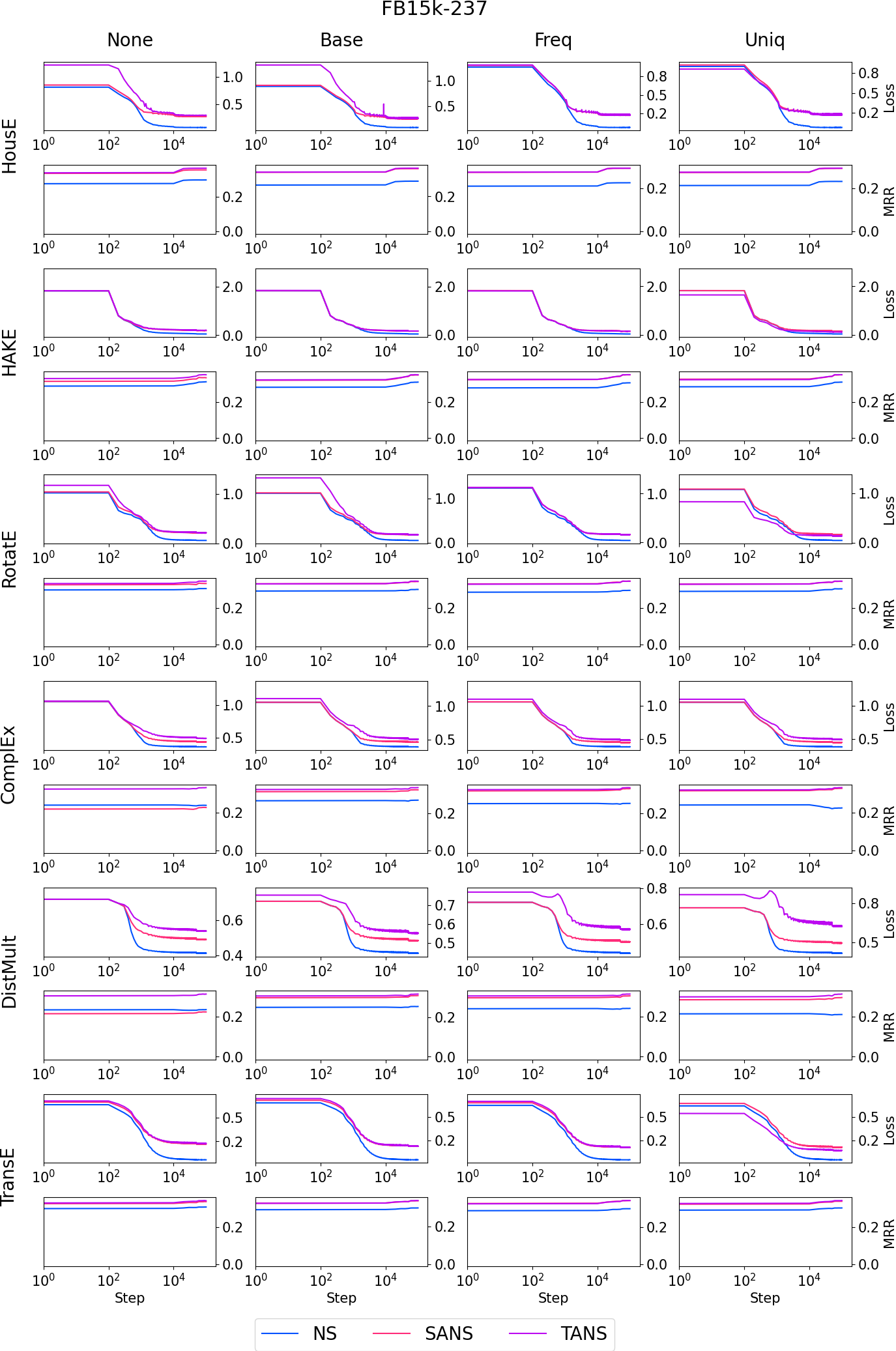}
    \caption{Training loss and validation MRR Curve on FB15k-237.}
    \label{fig:full_exp_train_valid_curves_fb}
\end{figure*}

\begin{figure*}[t]
    \centering
    \includegraphics[width=\textwidth]{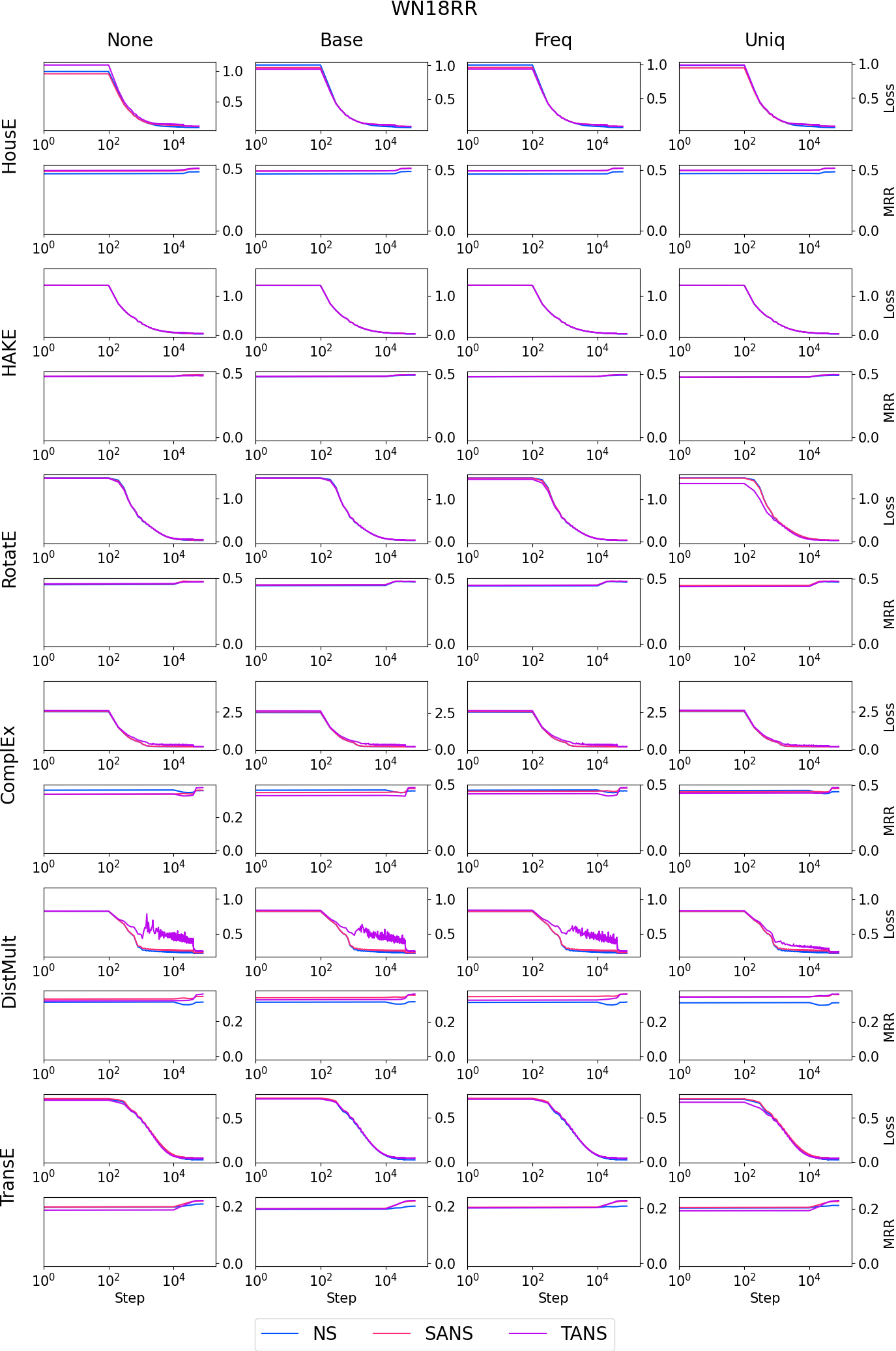}
    \caption{Training loss and validation MRR Curve on WN18RR.}
    \label{fig:full_exp_train_valid_curves_wn}
\end{figure*}

\begin{figure*}[t]
    \centering
    \includegraphics[width=\textwidth]{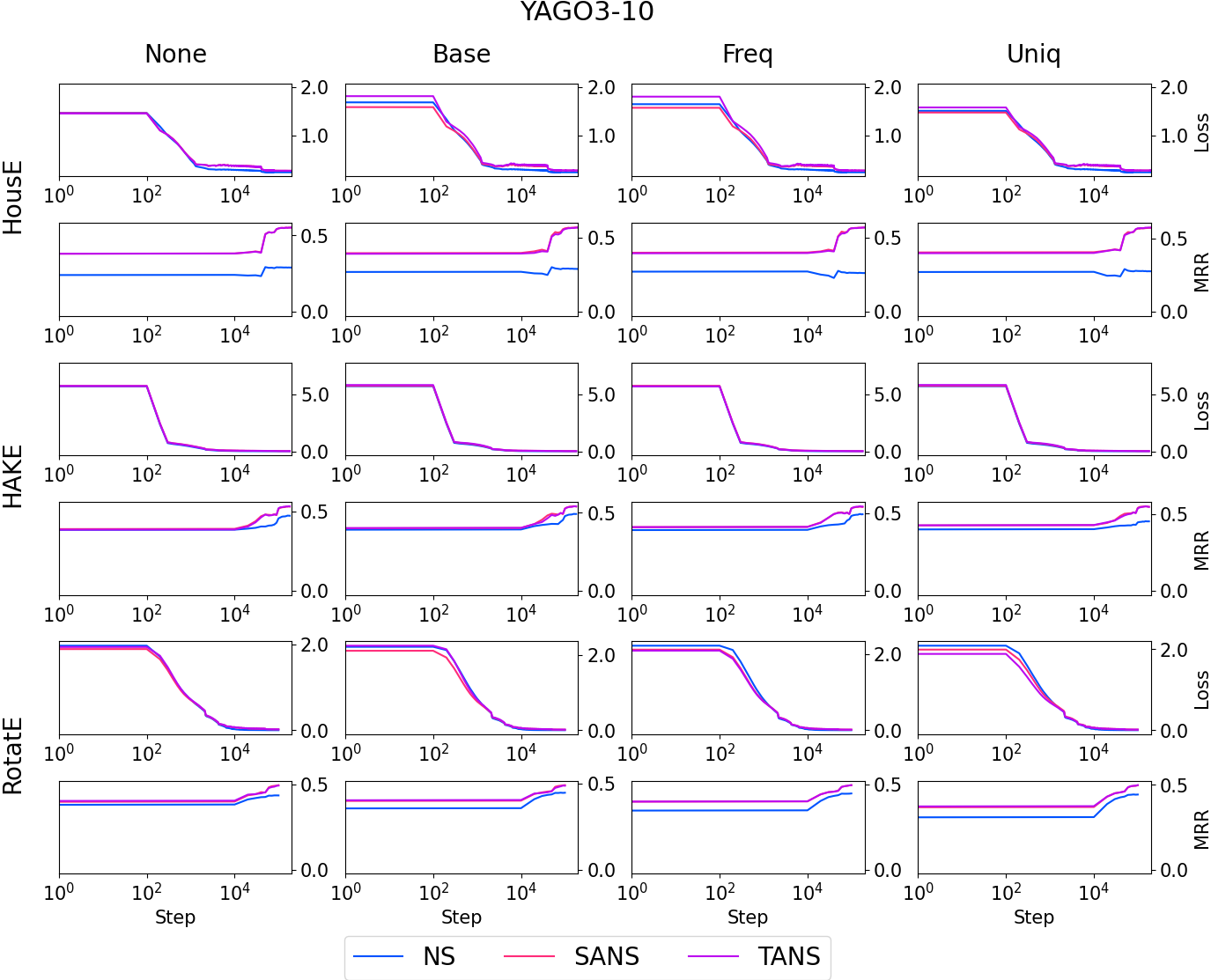}
    \caption{Training loss and validation MRR Curve on YAGO3-10.}
    \label{fig:full_exp_train_valid_curves_yago}
\end{figure*}

Figure~\ref{fig:full_exp_train_valid_curves_fb},~\ref{fig:full_exp_train_valid_curves_wn},~and~\ref{fig:full_exp_train_valid_curves_yago} show the training loss curves and validation MRR curves for each smoothing method. 
From these figures, we can understand that the convergence of TANS loss is as well as SANS and NS loss on datasets FB15k-237, WN18RR, and YAGO3-10 for each KGE model. Meanwhile, the time complexity of TANS is the same with SANS and NS loss too. 

\begin{figure*}[t]
    \centering
    \includegraphics[width=.93\textwidth]{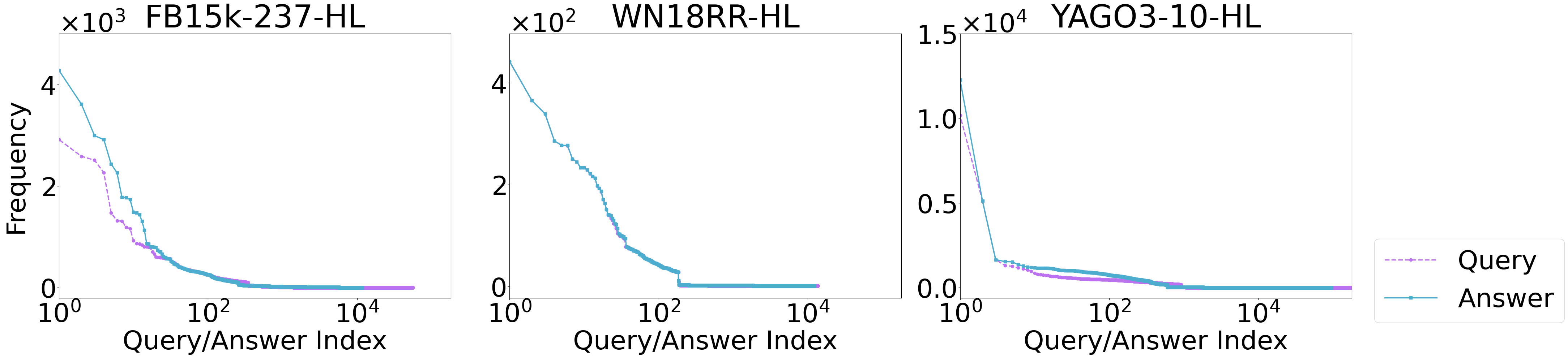}
    \caption{Appearance frequencies of queries and answers (entities) in the training data of the sparser subsets FB15k-237-HL, WN18RR-HL, and YAGO3-10-HL. Note that the indices are sorted from high frequency to low.}
    \label{fig:query_answer_freq_hl}
\end{figure*}
\begin{table*}[t]
    \centering
    \small
    \begin{minipage}{.49\textwidth}
    \resizebox{\textwidth}{!}{			
    \begin{tabular}{ccrrrr}
    \toprule
    Dataset & Split & \multicolumn{1}{l}{Tuple} & \multicolumn{1}{l}{Query} & \multicolumn{1}{l}{Entity} & \multicolumn{1}{l}{Relation} \\
    \midrule
    \multirow{4}[6]{*}{FB15k-237} & Total & 310,116 & 150,508 & 14,541 & 237 \\
\cmidrule{2-6}          & \#Train & 272,115 & 138,694 & 14,505 & 237 \\
\cmidrule{2-6}          & \#Valid & 17,535 & 19,750 & 9,809 & 223 \\
\cmidrule{2-6}          & \#Test & 20,466 & 22,379 & 10,348 & 224 \\
    \midrule
    \multirow{4}[6]{*}{WN18RR} & Total & 93,003 & 77,479 & 40,943 & 11 \\
\cmidrule{2-6}          & \#Train & 86,835 & 74,587 & 40,559 & 11 \\
\cmidrule{2-6}          & \#Valid & 3,034 & 5,431 & 5,173 & 11 \\
\cmidrule{2-6}          & \#Test & 3,134 & 5,565 & 5,323 & 11 \\
    \midrule
    \multirow{4}[6]{*}{YAGO3-10} & Total & 1,089,040 & 372,775 & 123,182 & 37 \\
\cmidrule{2-6}          & \#Train & 1,079,040 & 371,077 & 123,143 & 37 \\
\cmidrule{2-6}          & \#Valid & 5,000 & 8,534 & 7,948 & 33 \\
\cmidrule{2-6}          & \#Test & 5,000 & 8,531 & 7,937 & 34 \\
    \bottomrule
    \end{tabular}
    }
    \caption{Statistics for each public dataset.}
    \label{tab:common_datasets}
    \end{minipage}\hfill % \hfill 用于在两个 minipage 之间添加空格
    \begin{minipage}{.49\textwidth}
% \end{table}
% \begin{table}[t]
    \centering
    \small
    \resizebox{\textwidth}{!}{	
    \begin{tabular}{ccrrrr}
    \toprule
    Dataset & Split & \multicolumn{1}{l}{Tuple} & \multicolumn{1}{l}{Query} & \multicolumn{1}{l}{Entity} & \multicolumn{1}{l}{Relation} \\
    \midrule
    \multirow{4}[6]{*}{FB15k-237-HL} & Total & 111,631 & 63,330 & 11,828 & 155 \\
\cmidrule{2-6}          & \#Train & 95,244 & 55,923 & 11,600 & 155 \\
\cmidrule{2-6}          & \#Valid & 7,571 & 6,918 & 4,933 & 90 \\
\cmidrule{2-6}          & \#Test & 8,816 & 7,830 & 5,406 & 89 \\
    \midrule
    \multirow{4}[6]{*}{WN18RR-HL} & Total & 14,697 & 14,675 & 12,973 & 10 \\
\cmidrule{2-6}          & \#Train & 13,758 & 13,785 & 12,275 & 10 \\
\cmidrule{2-6}          & \#Valid & 465   & 619   & 613   & 9 \\
\cmidrule{2-6}          & \#Test & 474   & 623   & 619   & 8 \\
    \midrule
    \multirow{4}[6]{*}{YAGO3-10-HL} & Total & 366,079 & 182,274 & 95,788 & 29 \\
\cmidrule{2-6}          & \#Train & 362,728 & 181,196 & 95,432 & 29 \\
\cmidrule{2-6}          & \#Valid & 1,662 & 2,316 & 2,113 & 13 \\
\cmidrule{2-6}          & \#Test & 1,689 & 2,359 & 2,135 & 14 \\
    \bottomrule
    \end{tabular}
    }
    \caption{Statistics of the filtered sparser datasets.}
    \label{tab:artificial_datasets}
    \end{minipage}
\end{table*}

\section{Sparse Queries}
\subsection{Appearance Frequencies of Queries and Answers}
\label{fig:frequencies_hl}
Figure~\ref{fig:query_answer_freq_hl} shows the appearance frequencies of queries and answers in the training set of our filtered sparser data FB15k-237-HL, WN18RR-HL, and YAGO3-10-HL, expained in \S\ref{sec:analysis}.

\subsection{Data Statistics}
\label{app:subsec:sparse_stats}
Table~\ref{tab:artificial_datasets} shows detailed statistics of our filtered sparser data FB15k-237-HL, WN18RR-HL, and YAGO3-10-HL, expained in \S\ref{sec:analysis}.

\subsection{Detailed Results}
\label{app:subsec:sparse_results}
Table~\ref{tab:results_hl_fb}, ~\ref{tab:results_hl_wn}, and ~\ref{tab:results_hl_yago} shows the detailed results on our filtered sparser data FB15k-237-HL, WN18RR-HL, and YAGO3-10-HL, expained in \S\ref{sec:analysis}. Notations are as those described in \S\ref{app:sec:full_exp_table}.

\begin{table*}[htbp]
    \centering
    \resizebox{0.63\textwidth}{!}{
    \small
    {
    \renewcommand{\arraystretch}{0.415}
    \begin{tabular}{cccccccccccc}
    \toprule
    \multicolumn{12}{c}{FB15k-237} \\
    \midrule
    \multirow{2}[4]{*}{Model} & \multicolumn{2}{c}{Subsampling} & \multicolumn{2}{c}{MRR} & \multicolumn{2}{c}{H@1} & \multicolumn{2}{c}{H@3} & \multicolumn{2}{c}{H@10} & \multirow{1.5}[4]{*}{$\gamma$} \\
\cmidrule{2-11}          & Assumption & Loss  & Mean  & SD    & Mean  & SD    & Mean  & SD    & Mean  & SD    &  \\
    \midrule
    \multirow{15}[24]{*}{ComplEx} & \multirow{3}[6]{*}{None} & NS    & 23.9  & 0.2   & 15.8  & 0.1   & 26.1  & 0.3   & 40.0  & 0.2   & \textbf{-} \\
\cmidrule{3-12}          &       & SANS  & 22.3  & 0.1   & 13.8  & 0.1   & 24.2  & 0.0   & 39.5  & 0.2   & - \\
\cmidrule{3-12}          &       & TANS  & \textbf{32.8 } & 0.2   & \textbf{23.2 } & 0.1   & \textbf{36.2 } & 0.2   & \textbf{52.2 } & 0.1   & -2 \\
\cmidrule{2-12}          & \multirow{3}[6]{*}{Base} & NS    & 27.2  & 0.1   & 19.1  & 0.1   & 29.5  & 0.1   & 43.0  & 0.2   & - \\
\cmidrule{3-12}          &       & SANS  & 32.3  & 0.0   & 23.0  & 0.1   & 35.4  & 0.1   & 51.2  & 0.1   & - \\
\cmidrule{3-12}          &       & TANS  & $^{\dagger}${\textbf{33.3 }} & 0.0   & $^{\dagger}${\textbf{23.8 }} & 0.1   & $^{\dagger}${\textbf{36.9 }} & 0.1   & $^{\dagger}${\textbf{52.7 }} & 0.0   & -1 \\
\cmidrule{2-12}          & \multirow{3}[6]{*}{Freq} & NS    & 25.1  & 0.2   & 17.1  & 0.3   & 27.4  & 0.2   & 41.0  & 0.2   & - \\
\cmidrule{3-12}          &       & SANS  & 32.7  & 0.1   & 23.6  & 0.1   & 36.0  & 0.1   & 51.2  & 0.1   & - \\
\cmidrule{3-12}          &       & TANS  & $^{\dagger}${\textbf{33.3 }} & 0.0   & $^{\dagger}${\textbf{23.8 }} & 0.0   & \textbf{36.8 } & 0.1   & \textbf{52.1 } & 0.2   & -0.5 \\
\cmidrule{2-12}          & \multirow{3}[6]{*}{Uniq} & NS    & 22.8  & 0.4   & 14.7  & 0.5   & 24.7  & 0.4   & 39.0  & 0.1   & - \\
\cmidrule{3-12}          &       & SANS  & 32.6  & 0.0   & \textbf{23.5 } & 0.1   & 35.8  & 0.1   & 51.2  & 0.1   & - \\
\cmidrule{3-12}          &       & TANS  & \textbf{33.0 } & 0.1   & \textbf{23.5 } & 0.1   & \textbf{36.5 } & 0.1   & \textbf{52.1 } & 0.1   & -0.5 \\
    \midrule
    \multirow{15}[24]{*}{DistMult} & \multirow{3}[6]{*}{None} & NS    & 23.3  & 0.1   & 15.6  & 0.1   & 25.7  & 0.1   & 38.4  & 0.1   & - \\
\cmidrule{3-12}          &       & SANS  & 22.3  & 0.1   & 14.0  & 0.2   & 24.1  & 0.1   & 39.2  & 0.0   & - \\
\cmidrule{3-12}          &       & TANS  & \textbf{31.0 } & 0.1   & \textbf{21.7 } & 0.1   & \textbf{34.0 } & 0.1   & \textbf{49.6 } & 0.1   & -1 \\
\cmidrule{2-12}          & \multirow{3}[6]{*}{Base} & NS    & 25.4  & 0.1   & 17.9  & 0.1   & 27.6  & 0.1   & 40.4  & 0.1   & - \\
\cmidrule{3-12}          &       & SANS  & 30.8  & 0.1   & 21.9  & 0.1   & 33.6  & 0.1   & 48.4  & 0.1   & - \\
\cmidrule{3-12}          &       & TANS  & $^{\dagger}${\textbf{31.5 }} & 0.1   & $^{\dagger}${\textbf{22.4 }} & 0.1   & $^{\dagger}${\textbf{34.6 }} & 0.1   & $^{\dagger}${\textbf{49.7 }} & 0.0   & -0.5 \\
\cmidrule{2-12}          & \multirow{3}[6]{*}{Freq} & NS    & 24.0  & 0.1   & 16.7  & 0.2   & 25.9  & 0.1   & 38.4  & 0.1   & - \\
\cmidrule{3-12}          &       & SANS  & 29.9  & 0.0   & 21.2  & 0.1   & 32.8  & 0.0   & 47.5  & 0.1   & - \\
\cmidrule{3-12}          &       & TANS  & \textbf{30.7 } & 0.0   & \textbf{21.6 } & 0.0   & \textbf{34.0 } & 0.0   & \textbf{49.0 } & 0.0   & -1 \\
\cmidrule{2-12}          & \multirow{3}[6]{*}{Uniq} & NS    & 21.0  & 0.1   & 13.5  & 0.2   & 22.8  & 0.2   & 36.3  & 0.2   & - \\
\cmidrule{3-12}          &       & SANS  & 29.2  & 0.0   & 20.5  & 0.1   & 31.9  & 0.0   & 46.7  & 0.0   & - \\
\cmidrule{3-12}          &       & TANS  & \textbf{30.7 } & 0.1   & \textbf{21.5 } & 0.1   & \textbf{33.8 } & 0.1   & \textbf{49.3 } & 0.1   & -2 \\
    \midrule
    \multirow{15}[24]{*}{TransE} & \multirow{3}[6]{*}{None} & NS    & 30.4  & 0.0   & 21.3  & 0.1   & 33.4  & 0.1   & 48.5  & 0.0   & - \\
\cmidrule{3-12}          &       & SANS  & 33.0  & 0.1   & 22.9  & 0.1   & 37.2  & 0.1   & $^{\dagger}${\textbf{53.0 }} & 0.1   & - \\
\cmidrule{3-12}          &       & TANS  & \textbf{33.6 } & 0.0   & \textbf{23.9 } & 0.0   & \textbf{37.3 } & 0.0   & $^{\dagger}${\textbf{53.0 }} & 0.1   & -0.5 \\
\cmidrule{2-12}          & \multirow{3}[6]{*}{Base} & NS    & 29.4  & 0.1   & 20.0  & 0.1   & 32.8  & 0.0   & 48.1  & 0.0   & - \\
\cmidrule{3-12}          &       & SANS  & \textbf{33.0 } & 0.1   & 23.1  & 0.1   & \textbf{36.8 } & 0.1   & \textbf{52.7 } & 0.1   & - \\
\cmidrule{3-12}          &       & TANS  & \textbf{33.0 } & 0.0   & 23.1  & 0.0   & \textbf{36.8 } & 0.1   & \textbf{52.7 } & 0.1   & -0.1 \\
\cmidrule{2-12}          & \multirow{3}[6]{*}{Freq} & NS    & 29.3  & 0.1   & 20.0  & 0.1   & 32.8  & 0.1   & 47.8  & 0.1   & - \\
\cmidrule{3-12}          &       & SANS  & \textbf{33.5 } & 0.0   & \textbf{23.9 } & 0.1   & \textbf{37.2 } & 0.1   & \textbf{52.8 } & 0.1   & - \\
\cmidrule{3-12}          &       & TANS  & \textbf{33.5 } & 0.1   & \textbf{23.9 } & 0.1   & \textbf{37.2 } & 0.0   & \textbf{52.8 } & 0.1   & -0.1 \\
\cmidrule{2-12}          & \multirow{3}[6]{*}{Uniq} & NS    & 30.1  & 0.1   & 21.0  & 0.1   & 33.6  & 0.0   & 48.0  & 0.0   & - \\
\cmidrule{3-12}          &       & SANS  & 33.5  & 0.0   & 23.9  & 0.0   & 37.3  & 0.2   & 52.7  & 0.1   & - \\
\cmidrule{3-12}          &       & TANS  & $^{\dagger}${\textbf{34.0 }} & 0.1   & $^{\dagger}${\textbf{24.5 }} & 0.1   & $^{\dagger}${\textbf{37.7 }} & 0.1   & $^{\dagger}${\textbf{53.0 }} & 0.1   & 0.5 \\
    \midrule
    \multirow{15}[24]{*}{RotatE} & \multirow{3}[6]{*}{None} & NS    & 30.3  & 0.0   & 21.4  & 0.1   & 33.2  & 0.1   & 48.4  & 0.1   & - \\
\cmidrule{3-12}          &       & SANS  & 32.9  & 0.1   & 22.8  & 0.1   & 36.8  & 0.0   & 53.1  & 0.2   & - \\
\cmidrule{3-12}          &       & TANS  & \textbf{34.1 } & 0.1   & \textbf{24.6 } & 0.1   & \textbf{37.7 } & 0.1   & $^{\dagger}${\textbf{53.3 }} & 0.1   & -0.5 \\
\cmidrule{2-12}          & \multirow{3}[6]{*}{Base} & NS    & 29.5  & 0.0   & 20.3  & 0.0   & 32.7  & 0.1   & 47.9  & 0.0   & - \\
\cmidrule{3-12}          &       & SANS  & 33.6  & 0.1   & 23.9  & 0.1   & 37.3  & 0.1   & \textbf{53.1 } & 0.0   & - \\
\cmidrule{3-12}          &       & TANS  & \textbf{33.8 } & 0.0   & \textbf{24.2 } & 0.0   & \textbf{37.4 } & 0.0   & 53.0  & 0.1   & -0.5 \\
\cmidrule{2-12}          & \multirow{3}[6]{*}{Freq} & NS    & 29.4  & 0.1   & 20.2  & 0.1   & 32.6  & 0.1   & 47.6  & 0.1   & - \\
\cmidrule{3-12}          &       & SANS  & 34.0  & 0.1   & \textbf{24.6 } & 0.0   & \textbf{37.7 } & 0.0   & 53.0  & 0.0   & - \\
\cmidrule{3-12}          &       & TANS  & \textbf{34.1 } & 0.0   & \textbf{24.6 } & 0.0   & \textbf{37.7 } & 0.0   & \textbf{53.1 } & 0.1   & -0.01 \\
\cmidrule{2-12}          & \multirow{3}[6]{*}{Uniq} & NS    & 30.1  & 0.0   & 21.2  & 0.1   & 33.3  & 0.1   & 47.7  & 0.1   & - \\
\cmidrule{3-12}          &       & SANS  & 33.9  & 0.1   & 24.4  & 0.1   & 37.6  & 0.1   & 52.9  & 0.1   & - \\
\cmidrule{3-12}          &       & TANS  & $^{\dagger}${\textbf{34.2 }} & 0.0   & $^{\dagger}${\textbf{24.7 }} & 0.1   & $^{\dagger}${\textbf{37.8 }} & 0.0   & \textbf{53.1 } & 0.1   & 0.5 \\
    \midrule
    \multirow{15}[24]{*}{HAKE} & \multirow{3}[6]{*}{None} & NS    & 30.8  & 0.1   & 21.8  & 0.1   & 33.8  & 0.1   & 48.6  & 0.1   & - \\
\cmidrule{3-12}          &       & SANS  & 32.8  & 0.2   & 22.7  & 0.3   & 36.9  & 0.1   & 52.8  & 0.1   & - \\
\cmidrule{3-12}          &       & TANS  & \textbf{34.4 } & 0.1   & \textbf{24.9 } & 0.1   & \textbf{37.9 } & 0.2   & \textbf{53.6 } & 0.0   & -0.5 \\
\cmidrule{2-12}          & \multirow{3}[6]{*}{Base} & NS    & 30.4  & 0.1   & 21.6  & 0.1   & 33.3  & 0.1   & 48.2  & 0.0   & - \\
\cmidrule{3-12}          &       & SANS  & \textbf{34.1 } & 0.1   & \textbf{24.4 } & 0.1   & \textbf{37.9 } & 0.1   & 53.6  & 0.2   & - \\
\cmidrule{3-12}          &       & TANS  & \textbf{34.1 } & 0.0   & \textbf{24.4 } & 0.0   & \textbf{37.9 } & 0.0   & \textbf{53.7 } & 0.0   & -0.05 \\
\cmidrule{2-12}          & \multirow{3}[6]{*}{Freq} & NS    & 30.2  & 0.1   & 21.5  & 0.0   & 33.1  & 0.0   & 47.7  & 0.1   & - \\
\cmidrule{3-12}          &       & SANS  & \textbf{34.7 } & 0.0   & \textbf{25.2 } & 0.1   & \textbf{38.2 } & 0.0   & \textbf{53.8 } & 0.1   & - \\
\cmidrule{3-12}          &       & TANS  & 34.6  & 0.0   & 25.0  & 0.1   & \textbf{38.2 } & 0.2   & 53.7  & 0.1   & 0.05 \\
\cmidrule{2-12}          & \multirow{3}[6]{*}{Uniq} & NS    & 30.7  & 0.1   & 22.2  & 0.1   & 33.5  & 0.1   & 48.0  & 0.1   & - \\
\cmidrule{3-12}          &       & SANS  & 34.7  & 0.1   & 25.1  & 0.1   & 38.3  & 0.1   & 53.9  & 0.1   & - \\
\cmidrule{3-12}          &       & TANS  & $^{\dagger}${\textbf{34.9 }} & 0.0   & $^{\dagger}${\textbf{25.4 }} & 0.0   & $^{\dagger}${\textbf{38.6 }} & 0.1   & $^{\dagger}${\textbf{54.0 }} & 0.1   & 0.5 \\
    \midrule
    \multirow{15}[24]{*}{HousE} & \multirow{3}[6]{*}{None} & NS    & 29.1  & 0.1   & 20.6  & 0.1   & 31.6  & 0.1   & 46.3  & 0.1   & - \\
\cmidrule{3-12}          &       & SANS  & 34.7  & 0.2   & 24.8  & 0.2   & 38.5  & 0.3   & 54.4  & 0.2   & - \\
\cmidrule{3-12}          &       & TANS  & \textbf{35.6 } & 0.1   & \textbf{26.1 } & 0.1   & \textbf{39.4 } & 0.1   & \textbf{54.5 } & 0.1   & -1 \\
\cmidrule{2-12}          & \multirow{3}[6]{*}{Base} & NS    & 28.1  & 0.1   & 19.6  & 0.1   & 30.9  & 0.2   & 45.1  & 0.2   & - \\
\cmidrule{3-12}          &       & SANS  & 35.2  & 0.2   & 25.6  & 0.2   & 39.0  & 0.2   & 54.4  & 0.3   & - \\
\cmidrule{3-12}          &       & TANS  & \textbf{35.6 } & 0.1   & \textbf{26.1 } & 0.1   & \textbf{39.4 } & 0.2   & \textbf{54.5 } & 0.1   & -0.5 \\
\cmidrule{2-12}          & \multirow{3}[6]{*}{Freq} & NS    & 27.9  & 0.1   & 19.2  & 0.1   & 30.7  & 0.2   & 45.2  & 0.1   & - \\
\cmidrule{3-12}          &       & SANS  & \textbf{35.9 } & 0.2   & \textbf{26.4 } & 0.2   & 39.5  & 0.2   & \textbf{54.7 } & 0.1   & - \\
\cmidrule{3-12}          &       & TANS  & 35.8  & 0.2   & \textbf{26.4 } & 0.2   & \textbf{39.6 } & 0.2   & \textbf{54.7 } & 0.1   & -0.01 \\
\cmidrule{2-12}          & \multirow{3}[6]{*}{Uniq} & NS    & 28.8  & 0.1   & 20.2  & 0.2   & 31.9  & 0.1   & 45.7  & 0.0   & - \\
\cmidrule{3-12}          &       & SANS  & 36.1  & 0.1   & $^{\dagger}${\textbf{26.7 }} & 0.2   & 39.8  & 0.1   & $^{\dagger}${\textbf{54.8 }} & 0.2   & - \\
\cmidrule{3-12}          &       & TANS  & $^{\dagger}${\textbf{36.2 }} & 0.1   & $^{\dagger}${\textbf{26.7 }} & 0.2   & $^{\dagger}${\textbf{39.9 }} & 0.1   & $^{\dagger}${\textbf{54.8 }} & 0.1   & 0.1 \\
    \bottomrule
    \end{tabular}}
    }%
    \caption{Results on FB15k-237.\label{tab:fb15k-237}}
\end{table*}%

\begin{table*}[htbp]
    \centering
    \resizebox{0.59\textwidth}{!}{
    \small
    {
    \renewcommand{\arraystretch}{0.415}
    \begin{tabular}{cccccccccccc}
    \toprule
    \multicolumn{12}{c}{WN18RR} \\
    \midrule
    \multirow{2}[4]{*}{Model} & \multicolumn{2}{c}{Subsampling} & \multicolumn{2}{c}{MRR} & \multicolumn{2}{c}{H@1} & \multicolumn{2}{c}{H@3} & \multicolumn{2}{c}{H@10} & \multirow{1.5}[4]{*}{$\gamma$} \\
\cmidrule{2-11}          & Assumption & Loss  & Mean  & SD    & Mean  & SD    & Mean  & SD    & Mean  & SD    &  \\
    \midrule
    \multirow{15}[24]{*}{ComplEx} & \multirow{3}[6]{*}{None} & NS    & 44.5  & 0.1   & 38.1  & 0.2   & 48.3  & 0.2   & 55.5  & 0.1   & - \\
\cmidrule{3-12}          &       & SANS  & 45.0  & 0.1   & 41.0  & 0.1   & 46.5  & 0.3   & 53.3  & 0.3   & - \\
\cmidrule{3-12}          &       & TANS  & \textbf{47.3 } & 0.0   & \textbf{43.3 } & 0.0   & \textbf{49.1 } & 0.1   & \textbf{55.7 } & 0.1   & -2 \\
\cmidrule{2-12}          & \multirow{3}[6]{*}{Base} & NS    & 45.0  & 0.1   & 38.9  & 0.1   & 48.6  & 0.2   & 55.7  & 0.1   & - \\
\cmidrule{3-12}          &       & SANS  & 46.9  & 0.1   & 42.7  & 0.2   & 48.5  & 0.2   & 55.5  & 0.2   & - \\
\cmidrule{3-12}          &       & TANS  & \textbf{47.7 } & 0.2   & \textbf{43.6 } & 0.1   & \textbf{49.3 } & 0.2   & \textbf{55.9 } & 0.3   & -2 \\
\cmidrule{2-12}          & \multirow{3}[6]{*}{Freq} & NS    & 45.1  & 0.1   & 38.9  & 0.1   & 48.8  & 0.2   & 56.0  & 0.2   & - \\
\cmidrule{3-12}          &       & SANS  & 47.4  & 0.1   & 43.2  & 0.1   & 49.2  & 0.2   & 56.0  & 0.2   & - \\
\cmidrule{3-12}          &       & TANS  & \textbf{48.0 } & 0.1   & 43.9  & 0.1   & $^{\dagger}${\textbf{49.7 }} & 0.1   & \textbf{56.1 } & 0.1   & -2 \\
\cmidrule{2-12}          & \multirow{3}[6]{*}{Uniq} & NS    & 45.0  & 0.1   & 38.7  & 0.1   & 48.8  & 0.1   & 56.0  & 0.3   & - \\
\cmidrule{3-12}          &       & SANS  & 47.5  & 0.1   & 43.3  & 0.1   & 49.1  & 0.2   & 56.2  & 0.2   & - \\
\cmidrule{3-12}          &       & TANS  & $^{\dagger}${\textbf{48.3 }} & 0.1   & $^{\dagger}${\textbf{44.4 }} & 0.2   & 49.6  & 0.1   & $^{\dagger}${\textbf{56.3 }} & 0.2   & -1 \\
    \midrule
    \multirow{15}[24]{*}{DistMult} & \multirow{3}[6]{*}{None} & NS    & 38.5  & 0.2   & 30.6  & 0.3   & 42.9  & 0.2   & 52.5  & 0.1   & - \\
\cmidrule{3-12}          &       & SANS  & 42.4  & 0.0   & 38.2  & 0.1   & 43.7  & 0.0   & 51.0  & 0.2   & - \\
\cmidrule{3-12}          &       & TANS  & \textbf{44.2 } & 0.1   & \textbf{40.1 } & 0.1   & \textbf{45.3 } & 0.1   & \textbf{53.2 } & 0.2   & -2 \\
\cmidrule{2-12}          & \multirow{3}[6]{*}{Base} & NS    & 39.3  & 0.2   & 31.9  & 0.2   & 43.3  & 0.1   & 53.0  & 0.2   & - \\
\cmidrule{3-12}          &       & SANS  & 43.9  & 0.1   & 39.4  & 0.1   & 45.2  & 0.1   & 53.3  & 0.2   & - \\
\cmidrule{3-12}          &       & TANS  & \textbf{44.6 } & 0.0   & \textbf{40.5 } & 0.2   & \textbf{45.7 } & 0.1   & \textbf{53.9 } & 0.1   & -2 \\
\cmidrule{2-12}          & \multirow{3}[6]{*}{Freq} & NS    & 39.0  & 0.2   & 31.2  & 0.2   & 43.2  & 0.1   & 52.9  & 0.2   & - \\
\cmidrule{3-12}          &       & SANS  & 44.5  & 0.1   & 40.0  & 0.1   & \textbf{46.0 } & 0.1   & \textbf{54.2 } & 0.2   & - \\
\cmidrule{3-12}          &       & TANS  & \textbf{44.7 } & 0.1   & \textbf{40.5 } & 0.2   & 45.8  & 0.0   & 54.0  & 0.2   & -2 \\
\cmidrule{2-12}          & \multirow{3}[6]{*}{Uniq} & NS    & 38.8  & 0.2   & 30.8  & 0.2   & 43.1  & 0.1   & 53.0  & 0.2   & - \\
\cmidrule{3-12}          &       & SANS  & 44.7  & 0.1   & 40.1  & 0.1   & $^{\dagger}${\textbf{46.2 }} & 0.3   & 54.3  & 0.0   & - \\
\cmidrule{3-12}          &       & TANS  & $^{\dagger}${\textbf{45.0 }} & 0.1   & $^{\dagger}${\textbf{40.7 }} & 0.1   & 46.1  & 0.2   & $^{\dagger}${\textbf{54.5 }} & 0.2   & -0.5 \\
    \midrule
    \multirow{15}[24]{*}{TransE} & \multirow{3}[6]{*}{None} & NS    & 21.1  & 0.0   & 2.1   & 0.1   & 36.5  & 0.2   & 50.4  & 0.2   & - \\
\cmidrule{3-12}          &       & SANS  & 22.5  & 0.1   & 1.7   & 0.1   & \textbf{40.2 } & 0.1   & 52.5  & 0.2   & - \\
\cmidrule{3-12}          &       & TANS  & \textbf{22.7 } & 0.0   & \textbf{2.5 } & 0.0   & 39.5  & 0.2   & \textbf{53.4 } & 0.1   & 0.5 \\
\cmidrule{2-12}          & \multirow{3}[6]{*}{Base} & NS    & 20.3  & 0.1   & \textbf{1.6 } & 0.1   & 35.1  & 0.2   & 49.9  & 0.2   & - \\
\cmidrule{3-12}          &       & SANS  & 22.3  & 0.0   & 1.3   & 0.1   & \textbf{40.2 } & 0.1   & 52.9  & 0.1   & - \\
\cmidrule{3-12}          &       & TANS  & \textbf{22.4 } & 0.1   & 1.4   & 0.1   & 40.1  & 0.1   & \textbf{53.0 } & 0.1   & 0.1 \\
\cmidrule{2-12}          & \multirow{3}[6]{*}{Freq} & NS    & 21.0  & 0.1   & 1.8   & 0.1   & 36.4  & 0.2   & 51.0  & 0.2   & - \\
\cmidrule{3-12}          &       & SANS  & 23.0  & 0.0   & 1.9   & 0.1   & 40.9  & 0.2   & 53.6  & 0.0   & - \\
\cmidrule{3-12}          &       & TANS  & \textbf{23.1 } & 0.0   & \textbf{2.1 } & 0.0   & $^{\dagger}${\textbf{41.0 }} & 0.1   & \textbf{53.8 } & 0.0   & 0.1 \\
\cmidrule{2-12}          & \multirow{3}[6]{*}{Uniq} & NS    & 21.5  & 0.1   & 2.2   & 0.0   & 37.2  & 0.1   & 51.4  & 0.2   & - \\
\cmidrule{3-12}          &       & SANS  & 23.2  & 0.0   & 2.3   & 0.1   & \textbf{40.9 } & 0.2   & 53.6  & 0.1   & - \\
\cmidrule{3-12}          &       & TANS  & $^{\dagger}${\textbf{23.3 }} & 0.1   & $^{\dagger}${\textbf{3.0 }} & 0.0   & 40.2  & 0.2   & $^{\dagger}${\textbf{54.4 }} & 0.1   & 0.5 \\
    \midrule
    \multirow{15}[24]{*}{RotatE} & \multirow{3}[6]{*}{None} & NS    & 47.0  & 0.1   & 42.5  & 0.2   & 48.6  & 0.2   & 55.8  & 0.3   & - \\
\cmidrule{3-12}          &       & SANS  & \textbf{47.2 } & 0.1   & \textbf{42.6 } & 0.1   & \textbf{49.1 } & 0.1   & \textbf{56.7 } & 0.0   & - \\
\cmidrule{3-12}          &       & TANS  & \textbf{47.3 } & 0.1   & \textbf{42.6 } & 0.1   & \textbf{49.1 } & 0.1   & \textbf{56.7 } & 0.1   & -0.01 \\
\cmidrule{2-12}          & \multirow{3}[6]{*}{Base} & NS    & 47.0  & 0.0   & 42.2  & 0.1   & 48.7  & 0.1   & 56.3  & 0.1   & - \\
\cmidrule{3-12}          &       & SANS  & \textbf{47.5 } & 0.1   & \textbf{42.7 } & 0.2   & \textbf{49.3 } & 0.1   & \textbf{57.2 } & 0.1   & - \\
\cmidrule{3-12}          &       & TANS  & \textbf{47.5 } & 0.1   & \textbf{42.7 } & 0.2   & \textbf{49.3 } & 0.1   & 57.1  & 0.1   & 0.01 \\
\cmidrule{2-12}          & \multirow{3}[6]{*}{Freq} & NS    & 47.1  & 0.1   & 42.3  & 0.1   & 48.7  & 0.1   & 56.4  & 0.1   & - \\
\cmidrule{3-12}          &       & SANS  & \textbf{47.7 } & 0.1   & $^{\dagger}${\textbf{42.9 }} & 0.2   & 49.6  & 0.0   & \textbf{57.4 } & 0.1   & - \\
\cmidrule{3-12}          &       & TANS  & \textbf{47.7 } & 0.1   & 42.8  & 0.2   & \textbf{49.7 } & 0.1   & \textbf{57.4 } & 0.1   & 0.1 \\
\cmidrule{2-12}          & \multirow{3}[6]{*}{Uniq} & NS    & 47.2  & 0.2   & 42.7  & 0.2   & 48.7  & 0.1   & 56.3  & 0.1   & - \\
\cmidrule{3-12}          &       & SANS  & 47.7  & 0.1   & $^{\dagger}${\textbf{42.9 }} & 0.1   & 49.6  & 0.1   & 57.2  & 0.1   & - \\
\cmidrule{3-12}          &       & TANS  & $^{\dagger}${\textbf{47.8 }} & 0.2   & 42.8  & 0.3   & $^{\dagger}${\textbf{49.8 }} & 0.1   & $^{\dagger}${\textbf{57.6 }} & 0.1   & 0.5 \\
    \midrule
    \multirow{15}[24]{*}{HAKE} & \multirow{3}[6]{*}{None} & NS    & 48.8  & 0.1   & \textbf{44.5 } & 0.1   & 50.5  & 0.2   & 57.3  & 0.1   & - \\
\cmidrule{3-12}          &       & SANS  & \textbf{48.9 } & 0.0   & \textbf{44.5 } & 0.2   & \textbf{50.6 } & 0.3   & 57.7  & 0.1   & - \\
\cmidrule{3-12}          &       & TANS  & \textbf{48.9 } & 0.0   & 44.4  & 0.1   & 50.5  & 0.3   & \textbf{57.8 } & 0.1   & 0.01 \\
\cmidrule{2-12}          & \multirow{3}[6]{*}{Base} & NS    & 49.2  & 0.0   & 44.6  & 0.1   & 51.1  & 0.1   & 57.9  & 0.2   & - \\
\cmidrule{3-12}          &       & SANS  & \textbf{49.5 } & 0.1   & \textbf{45.0 } & 0.2   & 51.2  & 0.2   & 58.2  & 0.2   & - \\
\cmidrule{3-12}          &       & TANS  & \textbf{49.5 } & 0.1   & \textbf{45.0 } & 0.2   & 51.2  & 0.3   & \textbf{58.4 } & 0.2   & 0.1 \\
\cmidrule{2-12}          & \multirow{3}[6]{*}{Freq} & NS    & 49.3  & 0.1   & 44.8  & 0.1   & 51.3  & 0.2   & 58.0  & 0.2   & - \\
\cmidrule{3-12}          &       & SANS  & \textbf{49.7 } & 0.1   & \textbf{45.2 } & 0.2   & 51.5  & 0.1   & \textbf{58.4 } & 0.2   & - \\
\cmidrule{3-12}          &       & TANS  & \textbf{49.7 } & 0.0   & \textbf{45.2 } & 0.2   & \textbf{51.6 } & 0.3   & \textbf{58.4 } & 0.2   & -0.01 \\
\cmidrule{2-12}          & \multirow{3}[6]{*}{Uniq} & NS    & 49.4  & 0.2   & 44.9  & 0.2   & 51.3  & 0.2   & 57.8  & 0.2   & - \\
\cmidrule{3-12}          &       & SANS  & $^{\dagger}${\textbf{49.9 }} & 0.0   & 45.3  & 0.1   & $^{\dagger}${\textbf{51.8 }} & 0.2   & $^{\dagger}${\textbf{58.6 }} & 0.2   & - \\
\cmidrule{3-12}          &       & TANS  & $^{\dagger}${\textbf{49.9 }} & 0.1   & $^{\dagger}${\textbf{45.4 }} & 0.1   & $^{\dagger}${\textbf{51.8 }} & 0.2   & 58.5  & 0.2   & 0.05 \\
    \midrule
    \multirow{15}[24]{*}{HousE} & \multirow{3}[6]{*}{None} & NS    & 47.4  & 0.1   & 41.7  & 0.1   & 50.2  & 0.1   & 57.3  & 0.1   & - \\
\cmidrule{3-12}          &       & SANS  & 49.7  & 0.1   & 44.8  & 0.2   & 51.5  & 0.1   & 59.5  & 0.1   & - \\
\cmidrule{3-12}          &       & TANS  & \textbf{50.2 } & 0.1   & \textbf{45.3 } & 0.1   & \textbf{52.0 } & 0.1   & \textbf{60.0 } & 0.1   & -0.5 \\
\cmidrule{2-12}          & \multirow{3}[6]{*}{Base} & NS    & 48.1  & 0.1   & 42.4  & 0.1   & 50.9  & 0.1   & 58.5  & 0.2   & - \\
\cmidrule{3-12}          &       & SANS  & 51.2  & 0.1   & \textbf{46.7 } & 0.1   & \textbf{53.0 } & 0.2   & 60.3  & 0.1   & - \\
\cmidrule{3-12}          &       & TANS  & \textbf{51.3 } & 0.1   & \textbf{46.7 } & 0.2   & \textbf{53.0 } & 0.0   & \textbf{60.4 } & 0.1   & 0.05 \\
\cmidrule{2-12}          & \multirow{3}[6]{*}{Freq} & NS    & 48.1  & 0.2   & 42.5  & 0.3   & 50.9  & 0.2   & 58.5  & 0.2   & - \\
\cmidrule{3-12}          &       & SANS  & $^{\dagger}${\textbf{51.4 }} & 0.1   & $^{\dagger}${\textbf{46.8 }} & 0.1   & $^{\dagger}${\textbf{53.2 }} & 0.3   & $^{\dagger}${\textbf{60.5 }} & 0.1   & - \\
\cmidrule{3-12}          &       & TANS  & 51.3  & 0.2   & 46.7  & 0.2   & 53.1  & 0.3   & $^{\dagger}${\textbf{60.5 }} & 0.1   & 0.05 \\
\cmidrule{2-12}          & \multirow{3}[6]{*}{Uniq} & NS    & 48.1  & 0.1   & 42.5  & 0.1   & 50.8  & 0.2   & 58.1  & 0.1   & - \\
\cmidrule{3-12}          &       & SANS  & \textbf{51.2 } & 0.2   & $^{\dagger}${\textbf{46.8 }} & 0.2   & \textbf{52.7 } & 0.1   & \textbf{60.1 } & 0.1   & - \\
\cmidrule{3-12}          &       & TANS  & 51.1  & 0.3   & 46.7  & 0.5   & \textbf{52.7 } & 0.1   & 60.0  & 0.1   & -0.1 \\
    \bottomrule
    \end{tabular}}
    }%
    \caption{Results on WN18RR.\label{tab:wn18rr}}
\end{table*}%

\begin{table*}[htbp]
    \centering
    \resizebox{.9\textwidth}{!}{
    \small
    {
    \renewcommand{\arraystretch}{0.415}
    \begin{tabular}{cccccccccccc}
    \toprule
    \multicolumn{12}{c}{YAGO3-10} \\
    \midrule
    \multirow{2}[4]{*}{Model} & \multicolumn{2}{c}{Subsampling} & \multicolumn{2}{c}{MRR} & \multicolumn{2}{c}{H@1} & \multicolumn{2}{c}{H@3} & \multicolumn{2}{c}{H@10} & \multirow{1.5}[4]{*}{$\gamma$} \\
\cmidrule{2-11}          & Assumption & Loss  & Mean  & SD    & Mean  & SD    & Mean  & SD    & Mean  & SD    &  \\
    \midrule
    \multirow{15}[24]{*}{RotatE} & \multirow{3}[6]{*}{None} & NS    & 43.5  & 0.1   & 32.8  & 0.2   & 49.1  & 0.2   & 63.7  & 0.3   & - \\
\cmidrule{3-12}          &       & SANS  & \textbf{49.6 } & 0.2   & 39.9  & 0.1   & 55.3  & 0.3   & \textbf{67.3 } & 0.2   & - \\
\cmidrule{3-12}          &       & TANS  & \textbf{49.6 } & 0.2   & \textbf{40.0 } & 0.2   & \textbf{55.4 } & 0.5   & 67.2  & 0.3   & -0.05 \\
\cmidrule{2-12}          & \multirow{3}[6]{*}{Base} & NS    & 44.8  & 0.1   & 34.5  & 0.3   & 50.0  & 0.2   & 64.7  & 0.2   & - \\
\cmidrule{3-12}          &       & SANS  & \textbf{49.6 } & 0.3   & \textbf{40.1 } & 0.3   & \textbf{55.2 } & 0.4   & \textbf{67.4 } & 0.3   & - \\
\cmidrule{3-12}          &       & TANS  & 49.5  & 0.3   & \textbf{40.1 } & 0.3   & 55.0  & 0.5   & 67.3  & 0.3   & -0.05 \\
\cmidrule{2-12}          & \multirow{3}[6]{*}{Freq} & NS    & 44.8  & 0.2   & 34.5  & 0.3   & 50.0  & 0.1   & 64.7  & 0.2   & - \\
\cmidrule{3-12}          &       & SANS  & \textbf{49.9 } & 0.2   & \textbf{40.5 } & 0.3   & \textbf{55.5 } & 0.5   & \textbf{67.4 } & 0.3   & - \\
\cmidrule{3-12}          &       & TANS  & \textbf{49.9 } & 0.2   & \textbf{40.5 } & 0.3   & \textbf{55.5 } & 0.5   & \textbf{67.4 } & 0.2   & 0.01 \\
\cmidrule{2-12}          & \multirow{3}[6]{*}{Uniq} & NS    & 44.4  & 0.2   & 34.0  & 0.3   & 49.8  & 0.2   & 64.3  & 0.2   & - \\
\cmidrule{3-12}          &       & SANS  & 50.0  & 0.3   & 40.6  & 0.2   & 55.6  & 0.3   & 67.5  & 0.2   & - \\
\cmidrule{3-12}          &       & TANS  & $^{\dagger}${\textbf{50.1 }} & 0.2   & $^{\dagger}${\textbf{40.7 }} & 0.1   & $^{\dagger}${\textbf{55.7 }} & 0.3   & $^{\dagger}${\textbf{67.6 }} & 0.3   & 0.05 \\
    \midrule
    \multirow{15}[24]{*}{HAKE} & \multirow{3}[6]{*}{None} & NS    & 47.4  & 0.3   & 36.6  & 0.5   & 53.9  & 0.1   & 67.0  & 0.1   & - \\
\cmidrule{3-12}          &       & SANS  & 53.5  & 0.2   & 44.6  & 0.3   & \textbf{59.1 } & 0.4   & \textbf{69.0 } & 0.2   & - \\
\cmidrule{3-12}          &       & TANS  & \textbf{53.7 } & 0.1   & \textbf{45.3 } & 0.3   & 59.0  & 0.1   & 68.8  & 0.1   & 0.05 \\
\cmidrule{2-12}          & \multirow{3}[6]{*}{Base} & NS    & 48.8  & 0.3   & 38.4  & 0.4   & 55.0  & 0.2   & 68.1  & 0.3   & - \\
\cmidrule{3-12}          &       & SANS  & \textbf{54.6 } & 0.2   & \textbf{46.2 } & 0.3   & 59.9  & 0.2   & 69.6  & 0.2   & - \\
\cmidrule{3-12}          &       & TANS  & 54.5  & 0.2   & 45.9  & 0.3   & 59.9  & 0.2   & \textbf{69.9 } & 0.1   & -0.1 \\
\cmidrule{2-12}          & \multirow{3}[6]{*}{Freq} & NS    & 49.3  & 0.2   & 39.1  & 0.3   & 55.4  & 0.1   & 68.1  & 0.2   & - \\
\cmidrule{3-12}          &       & SANS  & 54.6  & 0.4   & 46.0  & 0.7   & \textbf{60.2 } & 0.1   & \textbf{69.6 } & 0.3   & - \\
\cmidrule{3-12}          &       & TANS  & \textbf{54.8 } & 0.2   & \textbf{46.4 } & 0.3   & 60.1  & 0.1   & \textbf{69.6 } & 0.3   & 0.05 \\
\cmidrule{2-12}          & \multirow{3}[6]{*}{Uniq} & NS    & 45.2  & 0.1   & 34.3  & 0.1   & 51.1  & 0.1   & 65.8  & 0.3   & - \\
\cmidrule{3-12}          &       & SANS  & $^{\dagger}${\textbf{55.2 }} & 0.3   & $^{\dagger}${\textbf{46.8 }} & 0.5   & $^{\dagger}${\textbf{60.5 }} & 0.2   & $^{\dagger}${\textbf{70.0 }} & 0.3   & - \\
\cmidrule{3-12}          &       & TANS  & 55.1  & 0.2   & $^{\dagger}${\textbf{46.8 }} & 0.3   & 60.3  & 0.1   & 69.9  & 0.2   & -0.1 \\
    \midrule
    \multirow{15}[24]{*}{HousE} & \multirow{3}[6]{*}{None} & NS    & 29.2  & 0.0   & 18.3  & 0.1   & 33.6  & 0.2   & 50.1  & 0.2   & - \\
\cmidrule{3-12}          &       & SANS  & \textbf{54.8 } & 1.3   & 46.8  & 1.3   & \textbf{59.7 } & 1.2   & \textbf{68.9 } & 1.2   & - \\
\cmidrule{3-12}          &       & TANS  & \textbf{54.8 } & 1.2   & \textbf{46.9 } & 1.2   & 59.6  & 1.2   & 68.8  & 1.1   & 0.01 \\
\cmidrule{2-12}          & \multirow{3}[6]{*}{Base} & NS    & 29.6  & 0.1   & 19.8  & 0.1   & 33.6  & 0.2   & 48.9  & 0.1   & - \\
\cmidrule{3-12}          &       & SANS  & 56.7  & 0.1   & 48.6  & 0.2   & 61.7  & 0.2   & 71.3  & 0.1   & - \\
\cmidrule{3-12}          &       & TANS  & \textbf{57.0 } & 0.2   & \textbf{49.0 } & 0.4   & \textbf{61.9 } & 0.3   & $^{\dagger}${\textbf{71.5 }} & 0.2   & -0.1 \\
\cmidrule{2-12}          & \multirow{3}[6]{*}{Freq} & NS    & 27.3  & 0.8   & 17.5  & 0.9   & 31.0  & 0.8   & 46.6  & 0.8   & - \\
\cmidrule{3-12}          &       & SANS  & 57.0  & 0.1   & 49.0  & 0.2   & 62.0  & 0.1   & \textbf{71.4 } & 0.1   & - \\
\cmidrule{3-12}          &       & TANS  & \textbf{57.2 } & 0.1   & \textbf{49.3 } & 0.1   & $^{\dagger}${\textbf{62.3 }} & 0.1   & \textbf{71.4 } & 0.1   & -0.1 \\
\cmidrule{2-12}          & \multirow{3}[6]{*}{Uniq} & NS    & 28.1  & 0.2   & 18.2  & 0.4   & 31.8  & 0.1   & 47.6  & 0.0   & - \\
\cmidrule{3-12}          &       & SANS  & 57.2  & 0.1   & 49.3  & 0.2   & 62.0  & 0.0   & 71.4  & 0.2   & - \\
\cmidrule{3-12}          &       & TANS  & $^{\dagger}${\textbf{57.3 }} & 0.2   & $^{\dagger}${\textbf{49.5 }} & 0.3   & \textbf{62.2 } & 0.1   & $^{\dagger}${\textbf{71.5 }} & 0.1   & -0.05 \\
    \bottomrule
    \end{tabular}}
    }%
    \caption{Results on YAGO3-10. \label{tab:YAGO3-10}}
\end{table*}%

\begin{table}[t]
    \centering
    \resizebox{.49\textwidth}{!}{		
    \begin{tabular}{cccccccccccccc}
    \toprule
    \multicolumn{8}{c}{FB15k-237-HL} \\
    \midrule
    \multirow{2}[4]{*}{Model} & \multicolumn{2}{c}{Subsampling} & \multicolumn{2}{c}{MRR} & \multicolumn{2}{c}{H@1} & \multirow{2}[4]{*}{$\gamma$} \\
\cmidrule{2-7}          & Assumption & Loss  & Mean  & SD    & Mean  & SD    &  \\
    \midrule
    \multirow{12}[24]{*}{HAKE} & \multirow{3}[6]{*}{None} & NS    & 38.1  & 0.3   & 28.4  & 0.5   & - \\
\cmidrule{3-8}          &       & SANS  & 35.2  & 0.2   & 24.5  & 0.3   & - \\
\cmidrule{3-8}          &       & TANS  & \textbf{41.1} & 0.1   & \textbf{33.0} & 0.1   & -1 \\
\cmidrule{2-8}          & \multirow{3}[6]{*}{Base} & NS    & 40.5  & 0.1   & 31.8  & 0.2   & - \\
\cmidrule{3-8}          &       & SANS  & 38.4  & 0.2   & 28.9  & 0.2   & - \\
\cmidrule{3-8}          &       & TANS  & \textbf{41.8} & 0.1   & \textbf{33.6} & 0.2   & -1 \\
\cmidrule{2-8}          & \multirow{3}[6]{*}{Freq} & NS    & 41.1  & 0.1   & 32.8  & 0.1   & - \\
\cmidrule{3-8}          &       & SANS  & 40.2  & 0.0   & 31.5  & 0.1   & - \\
\cmidrule{3-8}          &       & TANS  & $^{\dagger}${\textbf{42.0}} & 0.1   & $^{\dagger}${\textbf{33.7}} & 0.1   & -1 \\
\cmidrule{2-8}          & \multirow{3}[6]{*}{Uniq} & NS    & 41.5  & 0.1   & 33.2  & 0.1   & - \\
\cmidrule{3-8}          &       & SANS  & 41.1  & 0.0   & 32.8  & 0.0   & - \\
\cmidrule{3-8}          &       & TANS  & \textbf{41.9} & 0.2   & \textbf{33.5} & 0.2   & -0.1 \\
    \midrule
    \multirow{12}[24]{*}{RotatE} & \multirow{3}[6]{*}{None} & NS    & 40.0  & 0.1   & 30.8  & 0.1   & - \\
\cmidrule{3-8}          &       & SANS  & 36.3  & 0.1   & 25.3  & 0.2   & - \\
\cmidrule{3-8}          &       & TANS  & \textbf{41.5} & 0.0   & \textbf{33.1} & 0.1   & -1 \\
\cmidrule{2-8}          & \multirow{3}[6]{*}{Base} & NS    & 41.8  & 0.1   & 33.6  & 0.1   & - \\
\cmidrule{3-8}          &       & SANS  & 40.7  & 0.1   & 31.7  & 0.2   & - \\
\cmidrule{3-8}          &       & TANS  & \textbf{42.0} & 0.1   & \textbf{33.8} & 0.1   & -0.5 \\
\cmidrule{2-8}          & \multirow{3}[6]{*}{Freq} & NS    & 41.3  & 0.1   & 33.2  & 0.1   & - \\
\cmidrule{3-8}          &       & SANS  & 42.0  & 0.2   & 33.6  & 0.3   & - \\
\cmidrule{3-8}          &       & TANS  & $^{\dagger}${\textbf{42.3}} & 0.0   & $^{\dagger}${\textbf{34.1}} & 0.1   & -0.5 \\
\cmidrule{2-8}          & \multirow{3}[6]{*}{Uniq} & NS    & 41.7  & 0.1   & 33.7  & 0.2   & - \\
\cmidrule{3-8}          &       & SANS  & \textbf{42.2} & 0.1   & \textbf{33.8} & 0.2   & - \\
\cmidrule{3-8}          &       & TANS  & 42.1  & 0.1   & \textbf{33.8} & 0.2   & -0.05 \\
    \midrule
    \multirow{12}[24]{*}{HousE} & \multirow{3}[6]{*}{None} & NS    & 39.1  & 0.2   & 29.8  & 0.2   & - \\
\cmidrule{3-8}          &       & SANS  & 37.0  & 0.2   & 26.2  & 0.4   & - \\
\cmidrule{3-8}          &       & TANS  & \textbf{42.3} & 0.1   & \textbf{34.1} & 0.2   & -2 \\
\cmidrule{2-8}          & \multirow{3}[6]{*}{Base} & NS    & 40.3  & 0.1   & 31.3  & 0.2   & - \\
\cmidrule{3-8}          &       & SANS  & 40.5  & 0.4   & 31.3  & 0.4   & - \\
\cmidrule{3-8}          &       & TANS  & \textbf{42.4} & 0.2   & \textbf{34.2} & 0.3   & -2 \\
\cmidrule{2-8}          & \multirow{3}[6]{*}{Freq} & NS    & 39.8  & 0.3   & 31.0  & 0.3   & - \\
\cmidrule{3-8}          &       & SANS  & 42.1  & 0.2   & 33.8  & 0.2   & - \\
\cmidrule{3-8}          &       & TANS  & $^{\dagger}${\textbf{42.8}} & 0.3   & $^{\dagger}${\textbf{34.8}} & 0.4   & -1 \\
\cmidrule{2-8}          & \multirow{3}[6]{*}{Uniq} & NS    & 40.5  & 0.2   & 31.9  & 0.2   & - \\
\cmidrule{3-8}          &       & SANS  & 42.4  & 0.2   & 34.4  & 0.2   & - \\
\cmidrule{3-8}          &       & TANS  & \textbf{42.5} & 0.1   & \textbf{34.5} & 0.0   & -1 \\
    \bottomrule
    \end{tabular}}%
    \caption{Results on FB15k-237-HL.}
    \label{tab:results_hl_fb}%
\end{table}
\begin{table}[t]
    \centering
    \resizebox{.49\textwidth}{!}{		
    \begin{tabular}{cccccccccccccc}
    \toprule
    \multicolumn{8}{c}{WN18RR-HL} \\
    \midrule
    \multirow{2}[4]{*}{Model} & \multicolumn{2}{c}{Subsampling} & \multicolumn{2}{c}{MRR} & \multicolumn{2}{c}{H@1} & \multirow{2}[4]{*}{$\gamma$} \\
\cmidrule{2-7}          & Assumption & Loss  & Mean  & SD    & Mean  & SD    &  \\
    \midrule
    \multirow{12}[24]{*}{HAKE} & \multirow{3}[6]{*}{None} & NS    & 10.8  & 0.1   & 8.7   & 0.2   & - \\
\cmidrule{3-8}          &       & SANS  & 10.3  & 0.1   & 7.8   & 0.1   & - \\
\cmidrule{3-8}          &       & TANS  & \textbf{13.9} & 0.2   & $^{\dagger}${\textbf{12.1}} & 0.2   & -2 \\
\cmidrule{2-8}          & \multirow{3}[6]{*}{Base} & NS    & 12.1  & 0.2   & 9.5   & 0.3   & - \\
\cmidrule{3-8}          &       & SANS  & 11.1  & 0.1   & 9.1   & 0.1   & - \\
\cmidrule{3-8}          &       & TANS  & \textbf{13.7} & 0.1   & \textbf{11.7} & 0.3   & -2 \\
\cmidrule{2-8}          & \multirow{3}[6]{*}{Freq} & NS    & 12.4  & 0.1   & 10.4  & 0.1   & - \\
\cmidrule{3-8}          &       & SANS  & 11.9  & 0.2   & 9.5   & 0.2   & - \\
\cmidrule{3-8}          &       & TANS  & $^{\dagger}${\textbf{14.2}} & 0.5   & \textbf{11.9} & 0.4   & -2 \\
\cmidrule{2-8}          & \multirow{3}[6]{*}{Uniq} & NS    & 13.3  & 0.3   & 11.3  & 0.3   & - \\
\cmidrule{3-8}          &       & SANS  & 11.9  & 0.2   & 9.7   & 0.2   & - \\
\cmidrule{3-8}          &       & TANS  & \textbf{14.1} & 0.2   & \textbf{11.7} & 0.2   & -2 \\
    \midrule
    \multirow{12}[24]{*}{RotatE} & \multirow{3}[6]{*}{None} & NS    & 14.2  & 0.2   & \textbf{11.8} & 0.3   & - \\
\cmidrule{3-8}          &       & SANS  & 13.9  & 0.3   & 11.7  & 0.3   & - \\
\cmidrule{3-8}          &       & TANS  & \textbf{14.4} & 0.1   & \textbf{11.8} & 0.2   & -2 \\
\cmidrule{2-8}          & \multirow{3}[6]{*}{Base} & NS    & 13.9  & 0.2   & 11.5  & 0.2   & - \\
\cmidrule{3-8}          &       & SANS  & 14.1  & 0.3   & \textbf{11.7} & 0.3   & - \\
\cmidrule{3-8}          &       & TANS  & \textbf{14.5} & 0.1   & \textbf{11.7} & 0.1   & -2 \\
\cmidrule{2-8}          & \multirow{3}[6]{*}{Freq} & NS    & 14.4  & 0.1   & 12.0  & 0.1   & - \\
\cmidrule{3-8}          &       & SANS  & 14.3  & 0.4   & 12.0  & 0.3   & - \\
\cmidrule{3-8}          &       & TANS  & $^{\dagger}${\textbf{15.1}} & 0.1   & \textbf{12.2} & 0.1   & -2 \\
\cmidrule{2-8}          & \multirow{3}[6]{*}{Uniq} & NS    & 14.4  & 0.2   & 12.2  & 0.1   & - \\
\cmidrule{3-8}          &       & SANS  & 14.2  & 0.2   & 11.9  & 0.2   & - \\
\cmidrule{3-8}          &       & TANS  & $^{\dagger}${\textbf{15.1}} & 0.2   & $^{\dagger}${\textbf{12.3}} & 0.3   & -2 \\
    \midrule
    \multirow{12}[24]{*}{HousE} & \multirow{3}[6]{*}{None} & NS    & 10.7  & 1.8   & 8.4   & 1.4   & - \\
\cmidrule{3-8}          &       & SANS  & 11.7  & 1.1   & 9.5   & 0.9   & - \\
\cmidrule{3-8}          &       & TANS  & \textbf{13.4} & 0.4   & \textbf{11.0} & 0.4   & -2 \\
\cmidrule{2-8}          & \multirow{3}[6]{*}{Base} & NS    & 9.9   & 0.4   & 8.4   & 0.4   & - \\
\cmidrule{3-8}          &       & SANS  & 11.5  & 0.2   & 9.5   & 0.2   & - \\
\cmidrule{3-8}          &       & TANS  & \textbf{13.4} & 0.2   & \textbf{11.3} & 0.3   & -2 \\
\cmidrule{2-8}          & \multirow{3}[6]{*}{Freq} & NS    & $^{\dagger}${\textbf{13.9}} & 0.1   & 11.8  & 0.2   & - \\
\cmidrule{3-8}          &       & SANS  & 13.8  & 0.2   & 11.9  & 0.3   & - \\
\cmidrule{3-8}          &       & TANS  & $^{\dagger}${\textbf{13.9}} & 0.3   & $^{\dagger}${\textbf{12.0}} & 0.2   & 0.1 \\
\cmidrule{2-8}          & \multirow{3}[6]{*}{Uniq} & NS    & 13.7  & 0.1   & 11.6  & 0.1   & - \\
\cmidrule{3-8}          &       & SANS  & \textbf{13.8} & 0.2   & 11.6  & 0.2   & - \\
\cmidrule{3-8}          &       & TANS  & \textbf{13.8} & 0.2   & \textbf{11.7} & 0.3   & -0.05 \\
    \bottomrule
    \end{tabular}}%
    \caption{Results on WN18RR-HL.}
    \label{tab:results_hl_wn}%
\end{table}

\begin{table*}[t]
    \centering
    \resizebox{.6\textwidth}{!}{		
    \begin{tabular}{cccccccccccccc}
    \toprule
    \multicolumn{8}{c}{YAGO3-10-HL} \\
    \midrule
    \multirow{2}[4]{*}{Model} & \multicolumn{2}{c}{Subsampling} & \multicolumn{2}{c}{MRR} & \multicolumn{2}{c}{H@1} & \multirow{2}[4]{*}{$\gamma$} \\
\cmidrule{2-7}          & Assumption & Loss  & Mean  & SD    & Mean  & SD    &  \\
    \midrule
    \multirow{12}[24]{*}{HAKE} & \multirow{3}[6]{*}{None} & NS    & 45.9  & 0.0   & 36.9  & 0.1   & - \\
\cmidrule{3-8}          &       & SANS  & 47.8  & 0.4   & \textbf{40.0} & 0.6   & - \\
\cmidrule{3-8}          &       & TANS  & \textbf{49.2} & 0.4   & 39.8  & 0.7   & -0.5 \\
\cmidrule{2-8}          & \multirow{3}[6]{*}{Base} & NS    & \textbf{50.2} & 0.3   & \textbf{43.0} & 0.3   & - \\
\cmidrule{3-8}          &       & SANS  & 47.7  & 0.4   & 40.5  & 0.7   & - \\
\cmidrule{3-8}          &       & TANS  & 50.1  & 0.3   & 41.4  & 0.3   & -0.5 \\
\cmidrule{2-8}          & \multirow{3}[6]{*}{Freq} & NS    & $^{\dagger}${\textbf{50.8}} & 0.3   & $^{\dagger}${\textbf{43.3}} & 0.2   & - \\
\cmidrule{3-8}          &       & SANS  & 48.8  & 0.1   & 41.3  & 0.2   & - \\
\cmidrule{3-8}          &       & TANS  & 49.7  & 0.3   & 41.0  & 0.2   & -0.5 \\
\cmidrule{2-8}          & \multirow{3}[6]{*}{Uniq} & NS    & \textbf{49.4} & 0.2   & \textbf{40.8} & 0.2   & - \\
\cmidrule{3-8}          &       & SANS  & 46.9  & 0.4   & 39.8  & 0.5   & - \\
\cmidrule{3-8}          &       & TANS  & \textbf{49.4} & 0.6   & 40.6  & 0.8   & -0.5 \\
    \midrule
    \multirow{12}[24]{*}{RotatE} & \multirow{3}[6]{*}{None} & NS    & 38.0  & 0.1   & 28.7  & 0.3   & - \\
\cmidrule{3-8}          &       & SANS  & 41.3  & 0.1   & 32.3  & 0.2   & - \\
\cmidrule{3-8}          &       & TANS  & \textbf{43.5} & 0.1   & \textbf{34.8} & 0.2   & -0.5 \\
\cmidrule{2-8}          & \multirow{3}[6]{*}{Base} & NS    & 40.6  & 0.2   & 31.8  & 0.5   & - \\
\cmidrule{3-8}          &       & SANS  & \textbf{43.8} & 0.2   & 35.1  & 0.1   & - \\
\cmidrule{3-8}          &       & TANS  & \textbf{43.8} & 0.2   & \textbf{35.2} & 0.1   & -0.05 \\
\cmidrule{2-8}          & \multirow{3}[6]{*}{Freq} & NS    & 40.3  & 0.2   & 31.4  & 0.4   & - \\
\cmidrule{3-8}          &       & SANS  & 43.5  & 0.2   & 34.6  & 0.1   & - \\
\cmidrule{3-8}          &       & TANS  & \textbf{43.7} & 0.0   & \textbf{35.1} & 0.1   & -0.1 \\
\cmidrule{2-8}          & \multirow{3}[6]{*}{Uniq} & NS    & 40.2  & 0.0   & 31.3  & 0.2   & - \\
\cmidrule{3-8}          &       & SANS  & 43.9  & 0.1   & 35.1  & 0.2   & - \\
\cmidrule{3-8}          &       & TANS  & $^{\dagger}${\textbf{44.1}} & 0.1   & $^{\dagger}${\textbf{35.4}} & 0.3   & -0.1 \\
    \midrule
    \multirow{12}[24]{*}{HousE} & \multirow{3}[6]{*}{None} & NS    & 37.8  & 0.3   & 26.9  & 0.4   & - \\
\cmidrule{3-8}          &       & SANS  & 50.3  & 0.1   & 40.7  & 0.3   & - \\
\cmidrule{3-8}          &       & TANS  & $^{\dagger}${\textbf{52.5}} & 0.5   & $^{\dagger}${\textbf{45.4}} & 0.3   & -0.5 \\
\cmidrule{2-8}          & \multirow{3}[6]{*}{Base} & NS    & 42.8  & 1.2   & 34.3  & 1.9   & - \\
\cmidrule{3-8}          &       & SANS  & 51.9  & 0.3   & 44.4  & 0.2   & - \\
\cmidrule{3-8}          &       & TANS  & \textbf{51.9} & 0.6   & \textbf{44.3} & 0.8   & 0.05 \\
\cmidrule{2-8}          & \multirow{3}[6]{*}{Freq} & NS    & 39.7  & 0.8   & 29.9  & 1.5   & - \\
\cmidrule{3-8}          &       & SANS  & 48.6  & 1.7   & 40.0  & 1.4   & - \\
\cmidrule{3-8}          &       & TANS  & \textbf{52.0} & 0.1  & \textbf{44.5} & 0.3  & -1 \\
\cmidrule{2-8}          & \multirow{3}[6]{*}{Uniq} & NS    & 41.0  & 0.1   & 31.6  & 0.1   & - \\
\cmidrule{3-8}          &       & SANS  & 49.4  & 0.3   & 41.1  & 1.1   & - \\
\cmidrule{3-8}          &       & TANS  & \textbf{52.2} & 0.1   & \textbf{44.7} & 0.1   & -0.05 \\
    \bottomrule
    \end{tabular}}%
    \caption{Results on YAGO3-10-HL.}
    \label{tab:results_hl_yago}%
\end{table*}

\end{document}